\documentclass[submission,copyright,creativecommons]{eptcs}
\providecommand{\event}{QPL 2020} 
\usepackage{underscore}           

\usepackage{mathtools}
\usepackage{amsfonts}

\usepackage{amsmath,amsthm,amssymb}
\usepackage{xspace,enumerate,color,epsfig}
\usepackage{graphicx,float}

\usepackage{tikzit}
\usetikzlibrary{matrix}
\usetikzlibrary{fit}
\usetikzlibrary{calc}
\usepackage{tikz}
\usetikzlibrary{decorations.markings}
\usetikzlibrary{shapes.geometric}
\pgfdeclarelayer{edgelayer}
\pgfdeclarelayer{nodelayer}
\pgfsetlayers{edgelayer,nodelayer,main}

\tikzstyle{none}=[inner sep=0pt]
\definecolor{hexcolor0xff0000}{rgb}{1.000,0.000,0.000}
\definecolor{hexcolor0x000000}{rgb}{0.000,0.000,0.000}
\definecolor{hexcolor0x00ff00}{rgb}{0.000,1.000,0.000}
\definecolor{hexcolor0x000000}{rgb}{0.000,0.000,0.000}
\definecolor{hexcolor0xffff00}{rgb}{1.000,1.000,0.000}
\definecolor{hexcolor0xffffff}{rgb}{1.000,1.000,1.000}

\tikzstyle{rn}=[circle,fill=hexcolor0xff0000,draw=hexcolor0x000000,line width=0.8 pt]
\tikzstyle{gn}=[circle,fill=hexcolor0x00ff00,draw=hexcolor0x000000,line width=0.8 pt]
\tikzstyle{yn}=[circle,fill=hexcolor0xffff00,draw=hexcolor0x000000,line width=0.8 pt]
\tikzstyle{wn}=[circle,fill=hexcolor0xffffff,draw=hexcolor0x000000,line width=0.8 pt]
\tikzstyle{wnthick}=[circle,fill=hexcolor0xffffff,draw=hexcolor0x000000,line width=2.500]

\tikzstyle{simple}=[-,draw=hexcolor0x000000,line width=2.000]
\tikzstyle{arrow}=[-,draw=hexcolor0x000000,postaction={decorate},decoration={markings,mark=at position .5 with {\arrow{>}}},line width=2.000]
\tikzstyle{tick}=[-,draw=hexcolor0x000000,postaction={decorate},decoration={markings,mark=at position .5 with {\draw (0,-0.1) -- (0,0.1);}},line width=2.000]
\tikzstyle{halfthickness}=[-,draw=hexcolor0x000000,line width=0.500]
\tikzstyle{thick}=[-,draw=hexcolor0x000000,line width=2.500]
\tikzstyle{thicker}=[-,draw=hexcolor0x000000,line width=4.000]

\tikzstyle{env}=[copoint,regular polygon rotate=0,minimum width=0.2cm, fill=black]

\tikzstyle{probs}=[shape=semicircle,fill=white,draw=black,shape border rotate=180,minimum width=1.2cm]

\tikzstyle{every picture}=[baseline=-0.25em,scale=0.5]
\tikzstyle{dotpic}=[] 
\tikzstyle{diredges}=[every to/.style={diredge}]
\tikzstyle{math matrix}=[matrix of math nodes,left delimiter=(,right delimiter=),inner sep=2pt,column sep=1em,row sep=0.5em,nodes={inner sep=0pt},text height=1.5ex, text depth=0.25ex]

\tikzstyle{inline text}=[text height=1.5ex, text depth=0.25ex,yshift=0.5mm]
\tikzstyle{label}=[font=\footnotesize,text height=1.5ex, text depth=0.25ex,yshift=0.5mm]
\tikzstyle{left label}=[label,anchor=east,xshift=1.5mm]
\tikzstyle{right label}=[label,anchor=west,xshift=-1.5mm]

\tikzstyle{braceedge}=[decorate,decoration={brace,amplitude=2mm,raise=-1mm}]
\tikzstyle{small braceedge}=[decorate,decoration={brace,amplitude=1mm,raise=-1mm}]

\tikzstyle{doubled}=[line width=1.6pt] 
\tikzstyle{boldedge}=[doubled,shorten <=-0.17mm,shorten >=-0.17mm]
\tikzstyle{boldedgegray}=[doubled,gray,shorten <=-0.17mm,shorten >=-0.17mm]

\tikzstyle{semidoubled}=[line width=1.4pt] 
\tikzstyle{semiboldedgegray}=[semidoubled,gray,shorten <=-0.17mm,shorten >=-0.17mm]

\tikzstyle{boldedgedashed}=[very thick,dashed,shorten <=-0.17mm,shorten >=-0.17mm]
\tikzstyle{vboldedgedashed}=[doubled,dashed,shorten <=-0.17mm,shorten >=-0.17mm]
\tikzstyle{left hook arrow}=[left hook-latex]
\tikzstyle{right hook arrow}=[right hook-latex]
\tikzstyle{sembracket}=[line width=0.5pt,shorten <=-0.07mm,shorten >=-0.07mm]

\tikzstyle{causal edge}=[->,thick,gray]
\tikzstyle{causal nondir}=[thick,gray]
\tikzstyle{timeline}=[thick,gray, dashed]

\tikzstyle{cedge}=[<->,thick,gray!70!white]

\tikzstyle{empty diagram}=[draw=gray!40!white,dashed,shape=rectangle,minimum width=1cm,minimum height=1cm]
\tikzstyle{empty diagram small}=[draw=gray!50!white,dashed,shape=rectangle,minimum width=0.6cm,minimum height=0.5cm]

\tikzstyle{dot}=[inner sep=0mm,minimum width=2mm,minimum height=2mm,draw,shape=circle]
\tikzstyle{ddot}=[inner sep=0mm, doubled, minimum width=2.5mm,minimum height=2.5mm,draw,shape=circle]

\tikzstyle{black dot}=[dot,fill=black]
\tikzstyle{white dot}=[dot,fill=white,,text depth=-0.2mm]
\tikzstyle{green dot}=[white dot] 
\tikzstyle{gray dot}=[dot,fill=gray!40!white,,text depth=-0.2mm]

 \tikzstyle{red dot}=[dot,fill=red,font=\color{white}]

\tikzstyle{black ddot}=[ddot,fill=black]
\tikzstyle{white ddot}=[ddot,fill=white]
\tikzstyle{gray ddot}=[ddot,fill=gray!40!white]

\tikzstyle{gray edge}=[gray!40!white]

\tikzstyle{small dot}=[inner sep=0.5mm,minimum width=0pt,minimum height=0pt,draw,shape=circle]

\tikzstyle{small black dot}=[small dot,fill=black]
\tikzstyle{small white dot}=[small dot,fill=white]
\tikzstyle{small gray dot}=[small dot,fill=gray!40!white]

\tikzstyle{causal dot}=[inner sep=0.4mm,minimum width=0pt,minimum height=0pt,draw=white,shape=circle,fill=gray!40!white]

\tikzstyle{phase dimensions}=[minimum size=5mm,font=\footnotesize,rectangle,rounded corners=2.5mm,inner sep=0.2mm,outer sep=-2mm]
\tikzstyle{dphase dimensions}=[minimum size=5mm,font=\footnotesize,rectangle,rounded corners=2.5mm,inner sep=0.2mm,outer sep=-2mm]

\tikzstyle{white phase dot}=[dot,fill=white,phase dimensions]
\tikzstyle{white phase ddot}=[ddot,fill=white,dphase dimensions]

\tikzstyle{white rect ddot}=[draw=black,fill=white,doubled,minimum size=5mm,font=\footnotesize,rectangle,rounded corners=2.5mm,inner sep=0.2mm]
\tikzstyle{gray rect ddot}=[draw=black,fill=gray!40!white,doubled,minimum size=6mm,font=\footnotesize,rectangle,rounded corners=3mm]

\tikzstyle{gray phase dot}=[dot,fill=gray!40!white,phase dimensions]
\tikzstyle{gray phase ddot}=[ddot,fill=gray!40!white,dphase dimensions]
\tikzstyle{grey phase dot}=[gray phase dot]
\tikzstyle{grey phase ddot}=[gray phase ddot]

\tikzstyle{small phase dimensions}=[minimum size=4mm,font=\tiny,rectangle,rounded corners=2mm,inner sep=0.2mm,outer sep=-2mm]
\tikzstyle{small dphase dimensions}=[minimum size=4mm,font=\tiny,rectangle,rounded corners=2mm,inner sep=0.2mm,outer sep=-2mm]

\tikzstyle{small gray phase dot}=[dot,fill=gray!40!white,small phase dimensions]
\tikzstyle{small gray phase ddot}=[ddot,fill=gray!40!white,small dphase dimensions]

\tikzstyle{small map}=[draw,shape=rectangle,minimum height=4mm,minimum width=4mm,fill=white]

\tikzstyle{cnot}=[fill=white,shape=circle,inner sep=-1.4pt]

\tikzstyle{asym hadamard}=[fill=white,draw,shape=NEbox,inner sep=0.6mm,font=\footnotesize,minimum height=4mm]
\tikzstyle{asym hadamard conj}=[fill=white,draw,shape=NWbox,inner sep=0.6mm,font=\footnotesize,minimum height=4mm]
\tikzstyle{asym hadamard dag}=[fill=white,draw,shape=SEbox,inner sep=0.6mm,font=\footnotesize,minimum height=4mm]

\tikzstyle{hadamard}=[fill=white,draw,inner sep=0.6mm,font=\footnotesize,minimum height=4mm,minimum width=4mm]
\tikzstyle{small hadamard}=[fill=white,draw,inner sep=0.6mm,minimum height=1.5mm,minimum width=1.5mm]
\tikzstyle{dhadamard}=[hadamard,doubled]
\tikzstyle{small dhadamard}=[small hadamard,doubled]
\tikzstyle{small dhadamard rotate}=[small hadamard,doubled,rotate=45]
\tikzstyle{antipode}=[white dot,inner sep=0.3mm,font=\footnotesize]

\tikzstyle{scalar}=[diamond,draw,inner sep=0.5pt,font=\small]
\tikzstyle{dscalar}=[diamond,doubled, draw,inner sep=0.5pt,font=\small]

\tikzstyle{small box}=[rectangle,inline text,fill=white,draw,minimum height=5mm,yshift=-0.5mm,minimum width=5mm,font=\small]
\tikzstyle{small gray box}=[small box,fill=gray!30]
\tikzstyle{medium box}=[rectangle,inline text,fill=white,draw,minimum height=5mm,yshift=-0.5mm,minimum width=10mm,font=\small]
\tikzstyle{square box}=[small box] 
\tikzstyle{medium gray box}=[small box,fill=gray!30]
\tikzstyle{semilarge box}=[rectangle,inline text,fill=white,draw,minimum height=5mm,yshift=-0.5mm,minimum width=12.5mm,font=\small]
\tikzstyle{large box}=[rectangle,inline text,fill=white,draw,minimum height=5mm,yshift=-0.5mm,minimum width=15mm,font=\small]
\tikzstyle{large gray box}=[small box,fill=gray!30]

\tikzstyle{Bayes box}=[rectangle,fill=black,draw, minimum height=3mm, minimum width=3mm]

\tikzstyle{gray square point}=[small box,fill=gray!50]

\tikzstyle{dphase box white}=[dhadamard]
\tikzstyle{dphase box gray}=[dhadamard,fill=gray!50!white]

\tikzstyle{point}=[regular polygon,regular polygon sides=3,draw,scale=0.75,inner sep=-0.5pt,minimum width=9mm,fill=white,regular polygon rotate=180]
\tikzstyle{copoint}=[regular polygon,regular polygon sides=3,draw,scale=0.75,inner sep=-0.5pt,minimum width=9mm,fill=white]
\tikzstyle{dpoint}=[point,doubled]
\tikzstyle{dcopoint}=[copoint,doubled]

\tikzstyle{wide copoint}=[fill=white,draw,shape=isosceles triangle,shape border rotate=90,isosceles triangle stretches=true,inner sep=0pt,minimum width=1.5cm,minimum height=6.12mm]
\tikzstyle{wide point}=[fill=white,draw,shape=isosceles triangle,shape border rotate=-90,isosceles triangle stretches=true,inner sep=0pt,minimum width=1.5cm,minimum height=6.12mm,yshift=-0.0mm]
\tikzstyle{wide point plus}=[fill=white,draw,shape=isosceles triangle,shape border rotate=-90,isosceles triangle stretches=true,inner sep=0pt,minimum width=1.74cm,minimum height=7mm,yshift=-0.0mm]

\tikzstyle{wide dpoint}=[fill=white,doubled,draw,shape=isosceles triangle,shape border rotate=-90,isosceles triangle stretches=true,inner sep=0pt,minimum width=1.5cm,minimum height=6.12mm,yshift=-0.0mm]
\tikzstyle{wide dcopoint}=[fill=white,doubled,draw,shape=isosceles triangle,shape border rotate=90,isosceles triangle stretches=true,inner sep=0pt,minimum width=1.5cm,minimum height=6.12mm,yshift=-0.0mm]

\tikzstyle{tinypoint}=[regular polygon,regular polygon sides=3,draw,scale=0.55,inner sep=-0.15pt,minimum width=6mm,fill=white,regular polygon rotate=180]

\tikzstyle{white point}=[point]
\tikzstyle{white dpoint}=[dpoint]
\tikzstyle{green point}=[white point] 
\tikzstyle{white copoint}=[copoint]
\tikzstyle{gray point}=[point,fill=gray!40!white]
\tikzstyle{gray dpoint}=[gray point,doubled]
\tikzstyle{red point}=[gray point] 
\tikzstyle{gray copoint}=[copoint,fill=gray!40!white]
\tikzstyle{gray dcopoint}=[gray copoint,doubled]

\tikzstyle{white point guide}=[regular polygon,regular polygon sides=3,font=\scriptsize,draw,scale=0.65,inner sep=-0.5pt,minimum width=9mm,fill=white,regular polygon rotate=180]

\tikzstyle{black point}=[point,fill=black,font=\color{white}]
\tikzstyle{black copoint}=[copoint,fill=black,font=\color{white}]

\tikzstyle{tiny gray point}=[tinypoint,fill=gray!40!white]

\tikzstyle{diredge}=[->]
\tikzstyle{ddiredge}=[<->]
\tikzstyle{rdiredge}=[<-]
\tikzstyle{thickdiredge}=[->, very thick]
\tikzstyle{pointer edge}=[->,very thick,gray]
\tikzstyle{pointer edge part}=[very thick,gray]
\tikzstyle{dashed edge}=[dashed]
\tikzstyle{thick dashed edge}=[very thick,dashed]
\tikzstyle{thick gray dashed edge}=[thick dashed edge,gray!40]
\tikzstyle{thick map edge}=[very thick,|->]

\makeatletter
\newcommand{\boxshape}[3]{%
\pgfdeclareshape{#1}{
\inheritsavedanchors[from=rectangle] 
\inheritanchorborder[from=rectangle]
\inheritanchor[from=rectangle]{center}
\inheritanchor[from=rectangle]{north}
\inheritanchor[from=rectangle]{south}
\inheritanchor[from=rectangle]{west}
\inheritanchor[from=rectangle]{east}
\backgroundpath{
\southwest \pgf@xa=\pgf@x \pgf@ya=\pgf@y
\northeast \pgf@xb=\pgf@x \pgf@yb=\pgf@y

\@tempdima=#2
\@tempdimb=#3

\pgfpathmoveto{\pgfpoint{\pgf@xa - 5pt + \@tempdima}{\pgf@ya}}
\pgfpathlineto{\pgfpoint{\pgf@xa - 5pt - \@tempdima}{\pgf@yb}}
\pgfpathlineto{\pgfpoint{\pgf@xb + 5pt + \@tempdimb}{\pgf@yb}}
\pgfpathlineto{\pgfpoint{\pgf@xb + 5pt - \@tempdimb}{\pgf@ya}}
\pgfpathlineto{\pgfpoint{\pgf@xa - 5pt + \@tempdima}{\pgf@ya}}
\pgfpathclose
}
}}

\boxshape{NEbox}{0pt}{5pt}
\boxshape{SEbox}{0pt}{-5pt}
\boxshape{NWbox}{5pt}{0pt}
\boxshape{SWbox}{-5pt}{0pt}
\boxshape{EBox}{-3pt}{3pt}
\boxshape{WBox}{3pt}{-3pt}
\makeatother

\tikzstyle{cloud}=[shape=cloud,draw,minimum width=1.5cm,minimum height=1.5cm]

\tikzstyle{map}=[draw,shape=NEbox,inner sep=2pt,minimum height=6mm,fill=white]
\tikzstyle{dashedmap}=[draw,dashed,shape=NEbox,inner sep=2pt,minimum height=6mm,fill=white]
\tikzstyle{mapdag}=[draw,shape=SEbox,inner sep=2pt,minimum height=6mm,fill=white]
\tikzstyle{mapadj}=[draw,shape=SEbox,inner sep=2pt,minimum height=6mm,fill=white]
\tikzstyle{maptrans}=[draw,shape=SWbox,inner sep=2pt,minimum height=6mm,fill=white]
\tikzstyle{mapconj}=[draw,shape=NWbox,inner sep=2pt,minimum height=6mm,fill=white]

\tikzstyle{medium map}=[draw,shape=NEbox,inner sep=2pt,minimum height=6mm,fill=white,minimum width=7mm]
\tikzstyle{medium map dag}=[draw,shape=SEbox,inner sep=2pt,minimum height=6mm,fill=white,minimum width=7mm]
\tikzstyle{medium map adj}=[draw,shape=SEbox,inner sep=2pt,minimum height=6mm,fill=white,minimum width=7mm]
\tikzstyle{medium map trans}=[draw,shape=SWbox,inner sep=2pt,minimum height=6mm,fill=white,minimum width=7mm]
\tikzstyle{medium map conj}=[draw,shape=NWbox,inner sep=2pt,minimum height=6mm,fill=white,minimum width=7mm]
\tikzstyle{semilarge map}=[draw,shape=NEbox,inner sep=2pt,minimum height=6mm,fill=white,minimum width=9.5mm]
\tikzstyle{semilarge map trans}=[draw,shape=SWbox,inner sep=2pt,minimum height=6mm,fill=white,minimum width=9.5mm]
\tikzstyle{semilarge map adj}=[draw,shape=SEbox,inner sep=2pt,minimum height=6mm,fill=white,minimum width=9.5mm]
\tikzstyle{semilarge map dag}=[draw,shape=SEbox,inner sep=2pt,minimum height=6mm,fill=white,minimum width=9.5mm]
\tikzstyle{semilarge map conj}=[draw,shape=NWbox,inner sep=2pt,minimum height=6mm,fill=white,minimum width=9.5mm]
\tikzstyle{large map}=[draw,shape=NEbox,inner sep=2pt,minimum height=6mm,fill=white,minimum width=12mm]
\tikzstyle{large map conj}=[draw,shape=NWbox,inner sep=2pt,minimum height=6mm,fill=white,minimum width=12mm]
\tikzstyle{very large map}=[draw,shape=NEbox,inner sep=2pt,minimum height=6mm,fill=white,minimum width=17mm]

\tikzstyle{medium dmap}=[draw,doubled,shape=NEbox,inner sep=2pt,minimum height=6mm,fill=white,minimum width=7mm]
\tikzstyle{medium dmap dag}=[draw,doubled,shape=SEbox,inner sep=2pt,minimum height=6mm,fill=white,minimum width=7mm]
\tikzstyle{medium dmap adj}=[draw,doubled,shape=SEbox,inner sep=2pt,minimum height=6mm,fill=white,minimum width=7mm]
\tikzstyle{medium dmap trans}=[draw,doubled,shape=SWbox,inner sep=2pt,minimum height=6mm,fill=white,minimum width=7mm]
\tikzstyle{medium dmap conj}=[draw,doubled,shape=NWbox,inner sep=2pt,minimum height=6mm,fill=white,minimum width=7mm]
\tikzstyle{semilarge dmap}=[draw,doubled,shape=NEbox,inner sep=2pt,minimum height=6mm,fill=white,minimum width=9.5mm]
\tikzstyle{semilarge dmap trans}=[draw,doubled,shape=SWbox,inner sep=2pt,minimum height=6mm,fill=white,minimum width=9.5mm]
\tikzstyle{semilarge dmap adj}=[draw,doubled,shape=SEbox,inner sep=2pt,minimum height=6mm,fill=white,minimum width=9.5mm]
\tikzstyle{semilarge dmap dag}=[draw,doubled,shape=SEbox,inner sep=2pt,minimum height=6mm,fill=white,minimum width=9.5mm]
\tikzstyle{semilarge dmap conj}=[draw,doubled,shape=NWbox,inner sep=2pt,minimum height=6mm,fill=white,minimum width=9.5mm]
\tikzstyle{large dmap}=[draw,doubled,shape=NEbox,inner sep=2pt,minimum height=6mm,fill=white,minimum width=12mm]
\tikzstyle{large dmap conj}=[draw,doubled,shape=NWbox,inner sep=2pt,minimum height=6mm,fill=white,minimum width=12mm]
\tikzstyle{large dmap trans}=[draw,doubled,shape=SWbox,inner sep=2pt,minimum height=6mm,fill=white,minimum width=12mm]
\tikzstyle{large dmap adj}=[draw,doubled,shape=SEbox,inner sep=2pt,minimum height=6mm,fill=white,minimum width=12mm]
\tikzstyle{large dmap dag}=[draw,doubled,shape=SEbox,inner sep=2pt,minimum height=6mm,fill=white,minimum width=12mm]
\tikzstyle{very large dmap}=[draw,doubled,shape=NEbox,inner sep=2pt,minimum height=6mm,fill=white,minimum width=19.5mm]

\tikzstyle{muxbox}=[draw,shape=rectangle,minimum height=3mm,minimum width=3mm,fill=white]
\tikzstyle{dmuxbox}=[muxbox,doubled]

\tikzstyle{box}=[draw,shape=rectangle,inner sep=2pt,minimum height=6mm,minimum width=6mm,fill=white]
\tikzstyle{dbox}=[draw,doubled,shape=rectangle,inner sep=2pt,minimum height=6mm,minimum width=6mm,fill=white]
\tikzstyle{dmap}=[draw,doubled,shape=NEbox,inner sep=2pt,minimum height=6mm,fill=white]
\tikzstyle{dmapdag}=[draw,doubled,shape=SEbox,inner sep=2pt,minimum height=6mm,fill=white]
\tikzstyle{dmapadj}=[draw,doubled,shape=SEbox,inner sep=2pt,minimum height=6mm,fill=white]
\tikzstyle{dmaptrans}=[draw,doubled,shape=SWbox,inner sep=2pt,minimum height=6mm,fill=white]
\tikzstyle{dmapconj}=[draw,doubled,shape=NWbox,inner sep=2pt,minimum height=6mm,fill=white]

\tikzstyle{ddmap}=[draw,doubled,dashed,shape=NEbox,inner sep=2pt,minimum height=6mm,fill=white]
\tikzstyle{ddmapdag}=[draw,doubled,dashed,shape=SEbox,inner sep=2pt,minimum height=6mm,fill=white]
\tikzstyle{ddmapadj}=[draw,doubled,dashed,shape=SEbox,inner sep=2pt,minimum height=6mm,fill=white]
\tikzstyle{ddmaptrans}=[draw,doubled,dashed,shape=SWbox,inner sep=2pt,minimum height=6mm,fill=white]
\tikzstyle{ddmapconj}=[draw,doubled,dashed,shape=NWbox,inner sep=2pt,minimum height=6mm,fill=white]

\boxshape{sNEbox}{0pt}{3pt}
\boxshape{sSEbox}{0pt}{-3pt}
\boxshape{sNWbox}{3pt}{0pt}
\boxshape{sSWbox}{-3pt}{0pt}
\tikzstyle{smap}=[draw,shape=sNEbox,fill=white]
\tikzstyle{smapdag}=[draw,shape=sSEbox,fill=white]
\tikzstyle{smapadj}=[draw,shape=sSEbox,fill=white]
\tikzstyle{smaptrans}=[draw,shape=sSWbox,fill=white]
\tikzstyle{smapconj}=[draw,shape=sNWbox,fill=white]

\tikzstyle{dsmap}=[draw,dashed,shape=sNEbox,fill=white]
\tikzstyle{dsmapdag}=[draw,dashed,shape=sSEbox,fill=white]
\tikzstyle{dsmaptrans}=[draw,dashed,shape=sSWbox,fill=white]
\tikzstyle{dsmapconj}=[draw,dashed,shape=sNWbox,fill=white]

\boxshape{mNEbox}{0pt}{10pt}
\boxshape{mSEbox}{0pt}{-10pt}
\boxshape{mNWbox}{10pt}{0pt}
\boxshape{mSWbox}{-10pt}{0pt}
\tikzstyle{mmap}=[draw,shape=mNEbox]
\tikzstyle{mmapdag}=[draw,shape=mSEbox]
\tikzstyle{mmaptrans}=[draw,shape=mSWbox]
\tikzstyle{mmapconj}=[draw,shape=mNWbox]

\tikzstyle{mmapgray}=[draw,fill=gray!40!white,shape=mNEbox]
\tikzstyle{smapgray}=[draw,fill=gray!40!white,shape=sNEbox]

\makeatletter
\pgfdeclareshape{cornerpoint}{
\inheritsavedanchors[from=rectangle] 
\inheritanchorborder[from=rectangle]
\inheritanchor[from=rectangle]{center}
\inheritanchor[from=rectangle]{north}
\inheritanchor[from=rectangle]{south}
\inheritanchor[from=rectangle]{west}
\inheritanchor[from=rectangle]{east}
\backgroundpath{
\southwest \pgf@xa=\pgf@x \pgf@ya=\pgf@y
\northeast \pgf@xb=\pgf@x \pgf@yb=\pgf@y

\pgfmathsetmacro{\pgf@shorten@left}{\pgfkeysvalueof{/tikz/shorten left}}
\pgfmathsetmacro{\pgf@shorten@right}{\pgfkeysvalueof{/tikz/shorten right}}

\pgfpathmoveto{\pgfpoint{0.5 * (\pgf@xa + \pgf@xb)}{\pgf@ya - 5pt}}
\pgfpathlineto{\pgfpoint{\pgf@xa - 8pt + \pgf@shorten@left}{\pgf@yb - 1.5 * \pgf@shorten@left}}
\pgfpathlineto{\pgfpoint{\pgf@xa - 8pt + \pgf@shorten@left}{\pgf@yb}}
\pgfpathlineto{\pgfpoint{\pgf@xb + 8pt - \pgf@shorten@right}{\pgf@yb}}
\pgfpathlineto{\pgfpoint{\pgf@xb + 8pt - \pgf@shorten@right}{\pgf@yb - 1.5 * \pgf@shorten@right}}
\pgfpathclose
}
}

\pgfdeclareshape{cornercopoint}{
\inheritsavedanchors[from=rectangle] 
\inheritanchorborder[from=rectangle]
\inheritanchor[from=rectangle]{center}
\inheritanchor[from=rectangle]{north}
\inheritanchor[from=rectangle]{south}
\inheritanchor[from=rectangle]{west}
\inheritanchor[from=rectangle]{east}
\backgroundpath{
\southwest \pgf@xa=\pgf@x \pgf@ya=\pgf@y
\northeast \pgf@xb=\pgf@x \pgf@yb=\pgf@y

\pgfmathsetmacro{\pgf@shorten@left}{\pgfkeysvalueof{/tikz/shorten left}}
\pgfmathsetmacro{\pgf@shorten@right}{\pgfkeysvalueof{/tikz/shorten right}}

\pgfpathmoveto{\pgfpoint{0.5 * (\pgf@xa + \pgf@xb)}{\pgf@yb + 5pt}}
\pgfpathlineto{\pgfpoint{\pgf@xa - 8pt + \pgf@shorten@left}{\pgf@ya + 1.5 * \pgf@shorten@left}}
\pgfpathlineto{\pgfpoint{\pgf@xa - 8pt + \pgf@shorten@left}{\pgf@ya}}
\pgfpathlineto{\pgfpoint{\pgf@xb + 8pt - \pgf@shorten@right}{\pgf@ya}}
\pgfpathlineto{\pgfpoint{\pgf@xb + 8pt - \pgf@shorten@right}{\pgf@ya + 1.5 * \pgf@shorten@right}}
\pgfpathclose
}
}

\pgfdeclareshape{langpoint}{
\inheritsavedanchors[from=rectangle] 
\inheritanchorborder[from=rectangle]
\inheritanchor[from=rectangle]{center}
\inheritanchor[from=rectangle]{north}
\inheritanchor[from=rectangle]{south}
\inheritanchor[from=rectangle]{west}
\inheritanchor[from=rectangle]{east}
\backgroundpath{
\southwest \pgf@xa=\pgf@x \pgf@ya=\pgf@y
\northeast \pgf@xb=\pgf@x \pgf@yb=\pgf@y

\pgfmathsetmacro{\pgf@shorten@left}{\pgfkeysvalueof{/tikz/shorten left}}
\pgfmathsetmacro{\pgf@shorten@right}{\pgfkeysvalueof{/tikz/shorten right}}

\pgfpathmoveto{\pgfpoint{0.5 * (\pgf@xa + \pgf@xb)}{\pgf@ya - 2pt}}
\pgfpathlineto{\pgfpoint{\pgf@xa - 8pt}{\pgf@yb - 3 * \pgf@shorten@left + 5pt}} 
\pgfpathlineto{\pgfpoint{\pgf@xa - 8pt}{\pgf@yb -1pt}}
\pgfpathlineto{\pgfpoint{\pgf@xb + 8pt}{\pgf@yb -1pt}}
\pgfpathlineto{\pgfpoint{\pgf@xb + 8pt}{\pgf@yb - 3 * \pgf@shorten@left + 5pt}}
\pgfpathclose
}
}

\pgfdeclareshape{langcopointhigh}{
\inheritsavedanchors[from=rectangle] 
\inheritanchorborder[from=rectangle]
\inheritanchor[from=rectangle]{center}
\inheritanchor[from=rectangle]{north}
\inheritanchor[from=rectangle]{south}
\inheritanchor[from=rectangle]{west}
\inheritanchor[from=rectangle]{east}
\backgroundpath{
\southwest \pgf@xa=\pgf@x \pgf@ya=\pgf@y
\northeast \pgf@xb=\pgf@x \pgf@yb=\pgf@y

\pgfmathsetmacro{\pgf@shorten@left}{\pgfkeysvalueof{/tikz/shorten left}}
\pgfmathsetmacro{\pgf@shorten@right}{\pgfkeysvalueof{/tikz/shorten right}}

\pgfpathmoveto{\pgfpoint{0.5 * (\pgf@xa + \pgf@xb)}{\pgf@yb +4pt}}
\pgfpathlineto{\pgfpoint{\pgf@xa - 8pt}{\pgf@ya + 3 * \pgf@shorten@left -1pt}} 
\pgfpathlineto{\pgfpoint{\pgf@xa - 8pt}{\pgf@ya + 1pt}}
\pgfpathlineto{\pgfpoint{\pgf@xb + 8pt}{\pgf@ya + 1pt}}
\pgfpathlineto{\pgfpoint{\pgf@xb + 8pt}{\pgf@ya + 3 * \pgf@shorten@left -1pt}}
\pgfpathclose
}
} 

\pgfdeclareshape{langcopoint}{
\inheritsavedanchors[from=rectangle] 
\inheritanchorborder[from=rectangle]
\inheritanchor[from=rectangle]{center}
\inheritanchor[from=rectangle]{north}
\inheritanchor[from=rectangle]{south}
\inheritanchor[from=rectangle]{west}
\inheritanchor[from=rectangle]{east}
\backgroundpath{
\southwest \pgf@xa=\pgf@x \pgf@ya=\pgf@y
\northeast \pgf@xb=\pgf@x \pgf@yb=\pgf@y

\pgfmathsetmacro{\pgf@shorten@left}{\pgfkeysvalueof{/tikz/shorten left}}
\pgfmathsetmacro{\pgf@shorten@right}{\pgfkeysvalueof{/tikz/shorten right}}

\pgfpathmoveto{\pgfpoint{0.5 * (\pgf@xa + \pgf@xb)}{\pgf@yb +0pt}}
\pgfpathlineto{\pgfpoint{\pgf@xa - 8pt}{\pgf@ya + 3 * \pgf@shorten@left - 5pt}} 
\pgfpathlineto{\pgfpoint{\pgf@xa - 8pt}{\pgf@ya + 1pt}}
\pgfpathlineto{\pgfpoint{\pgf@xb + 8pt}{\pgf@ya + 1pt}}
\pgfpathlineto{\pgfpoint{\pgf@xb + 8pt}{\pgf@ya + 3 * \pgf@shorten@left - 5pt}}
\pgfpathclose
}
}

\pgfdeclareshape{langrect}{
\inheritsavedanchors[from=rectangle] 
\inheritanchorborder[from=rectangle]
\inheritanchor[from=rectangle]{center}
\inheritanchor[from=rectangle]{north}
\inheritanchor[from=rectangle]{south}
\inheritanchor[from=rectangle]{west}
\inheritanchor[from=rectangle]{east}
\backgroundpath{
\southwest \pgf@xa=\pgf@x \pgf@ya=\pgf@y
\northeast \pgf@xb=\pgf@x \pgf@yb=\pgf@y

\pgfmathsetmacro{\pgf@shorten@left}{\pgfkeysvalueof{/tikz/shorten left}}
\pgfmathsetmacro{\pgf@shorten@right}{\pgfkeysvalueof{/tikz/shorten right}}

\pgfpathmoveto{\pgfpoint{\pgf@xa - 8pt}{\pgf@ya + 3 * \pgf@shorten@left - 5pt}} 
\pgfpathlineto{\pgfpoint{\pgf@xa - 8pt}{\pgf@ya + 1pt}}
\pgfpathlineto{\pgfpoint{\pgf@xb + 8pt}{\pgf@ya + 1pt}}
\pgfpathlineto{\pgfpoint{\pgf@xb + 8pt}{\pgf@ya + 3 * \pgf@shorten@left - 5pt}}
\pgfpathclose
}
}

\pgfdeclareshape{langrecthigh}{
\inheritsavedanchors[from=rectangle] 
\inheritanchorborder[from=rectangle]
\inheritanchor[from=rectangle]{center}
\inheritanchor[from=rectangle]{north}
\inheritanchor[from=rectangle]{south}
\inheritanchor[from=rectangle]{west}
\inheritanchor[from=rectangle]{east}
\backgroundpath{
\southwest \pgf@xa=\pgf@x \pgf@ya=\pgf@y
\northeast \pgf@xb=\pgf@x \pgf@yb=\pgf@y

\pgfmathsetmacro{\pgf@shorten@left}{\pgfkeysvalueof{/tikz/shorten left}}
\pgfmathsetmacro{\pgf@shorten@right}{\pgfkeysvalueof{/tikz/shorten right}}

\pgfpathmoveto{\pgfpoint{\pgf@xa - 8pt}{\pgf@ya + 3 * \pgf@shorten@left - 0pt}} 
\pgfpathlineto{\pgfpoint{\pgf@xa - 8pt}{\pgf@ya + 1pt}}
\pgfpathlineto{\pgfpoint{\pgf@xb + 8pt}{\pgf@ya + 1pt}}
\pgfpathlineto{\pgfpoint{\pgf@xb + 8pt}{\pgf@ya + 3 * \pgf@shorten@left - 0pt}}
\pgfpathclose
}
}

\makeatother

\pgfkeyssetvalue{/tikz/shorten left}{0pt}
\pgfkeyssetvalue{/tikz/shorten right}{0pt}

\tikzstyle{kpoint common}=[draw,fill=white,inner sep=1pt,minimum height=4mm]

\tikzstyle{langstate}=[shape=langcopoint,shorten left=5pt,kpoint common,font=\footnotesize]
\tikzstyle{langstatehigh}=[shape=langcopointhigh,shorten left=5pt,kpoint common,font=\footnotesize]
\tikzstyle{langeffect}=[shape=langpoint,shorten left=5pt,kpoint common,font=\footnotesize]
\tikzstyle{langbox}=[shape=langrect,shorten left=5pt,kpoint common,font=\footnotesize] 
\tikzstyle{langboxhigh}=[shape=langrecthigh,shorten left=5pt,kpoint common,font=\footnotesize] 

\tikzstyle{kpoint}=[shape=cornerpoint,shorten left=5pt,kpoint common]
\tikzstyle{kpoint adjoint}=[shape=cornercopoint,shorten left=5pt,kpoint common]

\tikzstyle{kpoint conjugate}=[shape=cornerpoint,shorten right=5pt,kpoint common]
\tikzstyle{kpoint transpose}=[shape=cornercopoint,shorten right=5pt,kpoint common]
\tikzstyle{kpoint symm}=[shape=cornerpoint,shorten left=5pt,shorten right=5pt,kpoint common]

\tikzstyle{black kpoint}=[shape=cornerpoint,shorten left=5pt,kpoint common,fill=black,font=\color{white}]
\tikzstyle{black kpoint adjoint}=[shape=cornercopoint,shorten left=5pt,kpoint common,fill=black,font=\color{white}]
\tikzstyle{black kpointadj}=[shape=cornercopoint,shorten left=5pt,kpoint common,fill=black,font=\color{white}]

\tikzstyle{black dkpoint}=[shape=cornerpoint,shorten left=5pt,kpoint common,fill=black, doubled,font=\color{white}]
\tikzstyle{black dkpoint adjoint}=[shape=cornercopoint,shorten left=5pt,kpoint common,fill=black, doubled,font=\color{white}]
\tikzstyle{black dkpointadj}=[shape=cornercopoint,shorten left=5pt,kpoint common,fill=black, doubled,font=\color{white}]

\tikzstyle{kpointdag}=[kpoint adjoint]
\tikzstyle{kpointadj}=[kpoint adjoint]
\tikzstyle{kpointconj}=[kpoint conjugate]
\tikzstyle{kpointtrans}=[kpoint transpose]

\tikzstyle{big kpoint}=[kpoint, minimum width=1.2 cm, minimum height=8mm, inner sep=4pt, text depth=3mm]

\tikzstyle{wide kpoint}=[kpoint, minimum width=1 cm, inner sep=2pt]
\tikzstyle{wide kpointdag}=[kpointdag, minimum width=1 cm, inner sep=2pt]
\tikzstyle{wide kpointconj}=[kpointconj, minimum width=1 cm, inner sep=2pt]
\tikzstyle{wide kpointtrans}=[kpointtrans, minimum width=1 cm, inner sep=2pt]

\tikzstyle{gray kpoint}=[kpoint,fill=gray!50!white]
\tikzstyle{gray kpointdag}=[kpointdag,fill=gray!50!white]
\tikzstyle{gray kpointadj}=[kpointadj,fill=gray!50!white]
\tikzstyle{gray kpointconj}=[kpointconj,fill=gray!50!white]
\tikzstyle{gray kpointtrans}=[kpointtrans,fill=gray!50!white]

\tikzstyle{gray dkpoint}=[kpoint,fill=gray!50!white,doubled]
\tikzstyle{gray dkpointdag}=[kpointdag,fill=gray!50!white,doubled]
\tikzstyle{gray dkpointadj}=[kpointadj,fill=gray!50!white,doubled]
\tikzstyle{gray dkpointconj}=[kpointconj,fill=gray!50!white,doubled]
\tikzstyle{gray dkpointtrans}=[kpointtrans,fill=gray!50!white,doubled]

\tikzstyle{white label}=[draw,fill=white,rectangle,inner sep=0.7 mm]
\tikzstyle{gray label}=[draw,fill=gray!50!white,rectangle,inner sep=0.7 mm]
\tikzstyle{black label}=[draw,fill=black,rectangle,inner sep=0.7 mm]

\tikzstyle{dkpoint}=[kpoint,doubled]
\tikzstyle{wide dkpoint}=[wide kpoint,doubled]
\tikzstyle{dkpointdag}=[kpoint adjoint,doubled]
\tikzstyle{wide dkpointdag}=[wide kpointdag,doubled]
\tikzstyle{dkcopoint}=[kpoint adjoint,doubled]
\tikzstyle{dkpointadj}=[kpoint adjoint,doubled]
\tikzstyle{dkpointconj}=[kpoint conjugate,doubled]
\tikzstyle{dkpointtrans}=[kpoint transpose,doubled]

\tikzstyle{kscalar}=[kpoint common, shape=EBox, inner xsep=-1pt, inner ysep=3pt,font=\small]
\tikzstyle{kscalarconj}=[kpoint common, shape=WBox, inner xsep=-1pt, inner ysep=3pt,font=\small]

 \tikzstyle{upground}=[circuit ee IEC,ground,rotate=90,scale=2.5]
 \tikzstyle{downground}=[circuit ee IEC,ground,rotate=-90,scale=2.5]
 \tikzstyle{bigground}=[regular polygon,regular polygon sides=3,draw=gray,scale=0.50,inner sep=-0.5pt,minimum width=10mm,fill=gray]

\tikzstyle{arrs}=[-latex,font=\small,auto]
\tikzstyle{arrow plain}=[arrs]
\tikzstyle{arrow dashed}=[dashed,arrs]
\tikzstyle{arrow bold}=[very thick,arrs]
\tikzstyle{arrow hide}=[draw=white!0,-]
\tikzstyle{arrow reverse}=[latex-]
\tikzstyle{cdnode}=[]

\tikzstyle{dot}=[inner sep=0.3mm, minimum width=2mm, minimum height=2mm, draw, shape=circle, font={\footnotesize}, tikzit fill=magenta]
\tikzstyle{white dot}=[dot, fill=white, text depth=-0.2mm, tikzit category=ZH-pf]
\tikzstyle{gray dot}=[dot, fill={rgb,255: red,170; green,170; blue,170}, text depth=-0.2mm, tikzit category=ZH-pf]
\tikzstyle{H box}=[rectangle, fill=white, draw=black, xscale=1, yscale=1, font={\footnotesize}, inner sep=0.75pt, minimum width=0.3cm, minimum height=0.3cm, scale=0.75, tikzit shape=rectangle]
\tikzstyle{ug}=[regular polygon, regular polygon sides=3, fill={zx_red}, draw=black, inner sep=0pt, minimum width=1em, tikzit draw=blue]
\tikzstyle{st}=[star, star points=5, fill=white, draw=black, inner sep=1.2pt, line width=1.2pt, tikzit fill=blue, tikzit draw=red, tikzit category=ZH-pf]
\tikzstyle{triangle}=[regular polygon, regular polygon sides=3, fill=white, draw=black, inner sep=0pt, minimum width=1em, tikzit draw=blue, tikzit category=ZH-pf]
\tikzstyle{not}=[fill={rgb,255: red,170; green,170; blue,170}, draw=black, shape=circle, dot, minimum width=2.5mm, label={center:\tiny$\bm\neg$}]
\tikzstyle{bbindex}=[font={\color{blue}\footnotesize}]
\tikzstyle{wide point}=[fill=white, draw, shape=isosceles triangle, shape border rotate=-90, isosceles triangle stretches=true, inner sep=0pt, minimum width=1.5cm, minimum height=6.12mm, yshift=-0.0mm]
\tikzstyle{medium gray box}=[semilarge box, fill={gray!30}]
\tikzstyle{small box}=[rectangle, inline text, fill=white, draw, minimum height=5mm, yshift=-0.5mm, minimum width=5mm, font={\small}]
\tikzstyle{small gray box}=[small box, fill={gray!30}]
\tikzstyle{medium box}=[rectangle, inline text, fill=white, draw, minimum height=5mm, yshift=-0.5mm, minimum width=8mm, font={\small}]

\tikzstyle{gray}=[-, draw={blue!60!white}, tikzit draw=blue]
\tikzstyle{brace edge}=[-, tikzit draw=blue, decorate, decoration={brace,amplitude=1mm,raise=-1mm}]
\tikzstyle{diredge}=[->]
\tikzstyle{bbox edge}=[-, draw={rgb,255: red,42; green,145; blue,255}]

\input{zh.tikzdefs}

\newtheoremstyle{myplainstyle}
  {2mm} 
  {4mm} 
  {} 
  {} 
  {\bfseries} 
  {.} 
  {.5em} 
  {} 

\theoremstyle{myplainstyle}
\newtheorem{theorem}{Theorem}[section]

\newtheorem{definition}[theorem]{Definition}

\newtheorem{example*}[theorem]{Example*}
\newtheorem{examples*}[theorem]{Examples*}
\newtheorem{remark}[theorem]{Remark}
\newtheorem{remark*}[theorem]{Remark*}

\newcommand{\suchthat}[2]{\left\{#1 \: \middle\vert \: #2\right\}}
\newcommand{\ket}[1]{\left|#1\right\rangle}
\newcommand{\bra}[1]{\left\langle#1\right|}

\title{Quantum Natural Language Processing\\
on Near-Term Quantum Computers}

\author{
    K. Meichanetzidis,
    G. De Felice,
    A. Toumi
    and
    B. Coecke
    \institute{Cambridge Quantum Computing}
    \institute{Department of Computer Science, University of Oxford}
    \email{\{konstantinos.meichanetzidis, giovanni.defelice, alexis.toumi, bob.coecke\}(at)cs.ox.ac.uk}
    \\
    \\
    S. Gogioso and
    N. Chiappori
    \institute{Hashberg}
    \email{\{stefano.gogioso, nicolo.chiappori\}(at)hashberg.io}
}

\def\titlerunning{QNLP on Near-Term Quantum Computers}
\def\authorrunning{K. Meichanetzidis et al.}

\begin{document}
\maketitle

\begin{abstract}
    In this work, we describe a full-stack pipeline for natural language processing on near-term quantum computers, aka QNLP.
    The language-modelling framework we employ is that of compositional distributional semantics (DisCoCat), which extends and complements the compositional structure of pregroup grammars.
    Within this model, the grammatical reduction of a sentence is interpreted as a diagram, encoding a specific interaction of words according to the grammar. It is this interaction which, together with a specific choice of word embedding, realises the meaning (or "semantics") of a sentence.
    Building on the formal quantum-like nature of such interactions, we present a method for mapping DisCoCat diagrams to quantum circuits. Our methodology is compatible both with NISQ devices and with established Quantum Machine Learning techniques, paving the way to near-term applications of quantum technology to natural language processing.
\end{abstract}

\section{Introduction}

In recent years, research has flourished in the rapidly emerging fields of quantum machine learning and quantum artificial intelligence \cite{Schuld2014,Dunjko2017,dunjko2018machine,peterwittekbook}.
These terms cover a vast range of topics, from consideration of agent-environment interaction in the quantum domain to potential gains in using quantum devices as subroutines for machine learning algorithms.
For the purposes of this work, quantum machine learning will refer to supervised or unsupervised machine learning employing \emph{variational quantum circuits} in place of deep neural networks \cite{Benedetti2019}.
The study of variational quantum circuits is important in itself, as they constitute the setting for quantum computational supremacy experiments in the current era of noisy intermediate scale quantum (NISQ) devices \cite{Preskill2018}.
In this work, we focus on natural language processing (NLP), a sub-field of machine learning covering a diverse interdisciplinary landscape.
Our contribution to the field will be the introduction of a framework for quantum natural language processing (QNLP), tailored for implementation on NISQ devices.

We consider a particular distributional-compositional-categorical model of meaning (DisCoCat) for natural language \cite{coecke2010mathematical}, mediating between the rule-based approaches to language syntax and the statistical approach to language semantics, most famously associated with John Rupert Firth's assertion that \textit{``You shall know a word by the company it keeps''}.
In DisCoCat, structure is introduced via a compositional grammar model, that of \emph{pregroup grammar}, which is then endowed with a ``distributional'' embedding of words into a vector space, where vector geometry captures the correlations between words according to some corpus.
The interplay between compositionality of the grammar and the distributional word representation gives rise to semantics for phrases and sentences: starting from embeddings for individual words (extracted from a corpus), the compositional structure of grammar makes it is possible to give meaning to larger syntactic units \cite{lambek2008word}.
Tasks such as phrase (or sentence) similarity or question answering can be then be cast into geometric questions about vectors and tensors, and solved computationally.

Compositional grammar models---such as pregroup grammars and context free grammars (CFG)---have a natural tensor structure \cite{gallego2017language,pestun2017tensor,Gogioso2016} and can be considered quantum-native \cite{Zeng2016,Arad2010,teleling}.
Building on the recent proposal of quantum algorithms for NLP task by Zeng and Coecke~\cite{Zeng2016}, we take advantage of the tensor structure in order to construct a map from DisCoCat models to variational quantum circuits, where ans\"atze corresponding to lexical categories---aka parts-of-speech (POS)---are connected according to the grammar to form circuits for arbitrary syntactic units.
In some of its applications, the original Zeng-Coecke algorithm relies on the existence of a quantum random access memory (QRAM) \cite{Giovannetti2008}, which is not yet known to be efficiently implementable in the absence of fault tolerant scalable quantum computers \cite{aaronson2015read,Ciliberto2018}.
Here we take a different approach, using the classical ans\"atz parameters to encode the distributional embedding and avoiding the need for QRAM entirely.
The cost function for the parameter optimisation is informed by a corpus, already parsed and POS-tagged by classical means.
Taken all together, the pipeline is as follows:

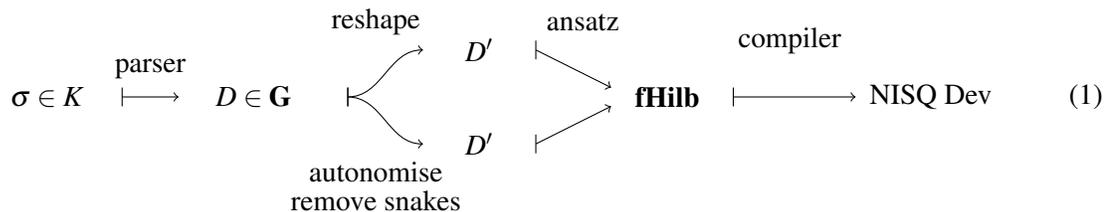
\begin{equation}
\begin{tikzpicture}
	\begin{pgfonlayer}{nodelayer}
		\node [style=none] (0) at (-10.75, 0) {$\sigma\in K$};
		\node [style=none] (1) at (-5.75, 0) {};
		\node [style=none] (2) at (-5.25, 0) {$D\in{\bf G}$};
		\node [style=none] (3) at (0.75, 1.25) {$D'$};
		\node [style=none] (5) at (12.75, 0) {NISQ Dev};
		\node [style=none] (6) at (-8.75, 0) {};
		\node [style=none] (7) at (-7.25, 0) {};
		\node [style=none] (8) at (-8.75, 0.25) {};
		\node [style=none] (9) at (-8.75, -0.25) {};
		\node [style=none] (10) at (7.5, 0) {};
		\node [style=none] (11) at (10.75, 0) {};
		\node [style=none] (12) at (7.5, 0.25) {};
		\node [style=none] (13) at (7.5, -0.25) {};
		\node [style=none] (14) at (2.25, 1.25) {};
		\node [style=none] (15) at (4.25, 0.25) {};
		\node [style=none] (16) at (2.25, 1.5) {};
		\node [style=none] (17) at (2.25, 1) {};
		\node [style=none] (18) at (-2.75, 0) {};
		\node [style=none] (19) at (-0.75, 1.25) {};
		\node [style=none] (20) at (-2.75, 0.25) {};
		\node [style=none] (21) at (-2.75, -0.25) {};
		\node [style=none] (22) at (-8, 0.75) {parser};
		\node [style=none] (23) at (3.5, 2) {ansatz};
		\node [style=none] (24) at (9, 1.5) {compiler};
		\node [style=none] (25) at (-2, 2) {reshape};
		\node [style=none] (26) at (-2.75, 0) {};
		\node [style=none] (27) at (-0.75, -1.25) {};
		\node [style=none] (28) at (-2.75, 0.25) {};
		\node [style=none] (29) at (-2.75, -0.25) {};
		\node [style=none] (30) at (-2, -2) {autonomise};
		\node [style=none] (31) at (0.75, -1.25) {$D'$};
		\node [style=none] (32) at (2.25, -1.25) {};
		\node [style=none] (33) at (4.25, -0.25) {};
		\node [style=none] (34) at (2.25, -1) {};
		\node [style=none] (35) at (2.25, -1.5) {};
		\node [style=none] (37) at (5.75, 0) {{\bf fHilb}};
		\node [style=none] (43) at (-2, -2.75) {remove snakes};
	\end{pgfonlayer}
	\begin{pgfonlayer}{edgelayer}
		\draw [style=diredge] (6.center) to (7.center);
		\draw (8.center) to (9.center);
		\draw [style=diredge, in=180, out=0, looseness=0.75] (10.center) to (11.center);
		\draw (12.center) to (13.center);
		\draw [style=diredge] (14.center) to (15.center);
		\draw (16.center) to (17.center);
		\draw [style=diredge, in=-180, out=0] (18.center) to (19.center);
		\draw (20.center) to (21.center);
		\draw [style=diredge, in=-180, out=0] (26.center) to (27.center);
		\draw (28.center) to (29.center);
		\draw [style=diredge] (32.center) to (33.center);
		\draw (34.center) to (35.center);
	\end{pgfonlayer}
\end{tikzpicture}
\end{equation}

\noindent
Firstly, a POS-tagged sentence $\sigma$ in a corpus $K$ is parsed to a diagram $D$ capturing its grammatical structure.
POS-tagging of sentences can be done by tools such as NLTK~\cite{nltk} or SpaCy~\cite{spacy}.
Furthermore, the POS-tags can be assigned pregroup types and pregroup parsing is proven to be efficient \cite{preller}.
Therefore, in this work,
we assume that the diagram obtained from a sentence is the input to our pipeline.
The diagram then
is further simplified to some other diagram $D'$.
In the case of the original non-NISQ-friendly
quantum compositional NLP of Zeng and Coecke \cite{Zeng2016},
this step is essential to retain the quantum advantage.
Here, we refine and improve on this method
as well as address unusual pregroup parsings with cycles.
We will also present an alternative method for simplifying a diagram.
The simplified diagram is then turned into a variational quantum circuit, which is finally compiled for NISQ devices.

Now, given a perfect compiler, the diagram simplification step can be bypassed.
On the other hand, diagram simplification is needed when
a simple transpiler is used.
Therefore, we consider all cases in this work.
Even though a perfect compiler is too much to ask for,
since usually NP-hard optimisation problems come into play,
one can do pretty well by using
state-of-the art software such as CQC's $\mathrm{t|ket\rangle}$,
a platform-agnostic compiler which interfaces with current NISQ architectures~\cite{sivarajah2020tketrangle}.
A Python wrapper for can be found at
{\tt github.com/CQCL/pytket}.

Now, regarding the mapping from sentence parsing diagrams to quantum circuits, the motivation is as following. 
Beyond the fact that variational quantum circuits are amenable to implementation on existing NISQ hardware, a reason for constructing such variational embeddings is to exploit an entirely novel feature space in which to encode the distributional semantics \cite{Havlicek2019}.
Quantum-enhanced feature spaces provide a dimension which is exponential in the number of qubits, so that QNLP models have the potential to take advantage of the space for data-intensive tasks.
Furthermore, the optimisation landscapes spanned by the variational quantum circuits are of different shape than those appearing in artificial neural networks, so there is the possibility for alternate performance profiles over equivalent benchmark tasks.

\section*{Contributions}

This work was originally commissioned by Cambridge Quantum Computing Ltd. (CQC) and was carried out independently by the CQC and Hashberg research groups.

\section{From Sentence to Diagram}

For the purposes of constructing our QNLP model, we build on work which uses \emph{pregroup grammars}, but context-free grammars (CFS) would be equally suitable for our construction. Note that pregroup grammars are weakly equivalent to CFGs \cite{weakequiv}.
To construct DisCoCat models of meaning, diagrams encoding the pregroup grammatical structure are constructed directly inside of compact closed categories---a special case of rigid categories---giving semantics to the model.

Specifically, the diagrams in this work represent complex matrices, i.e. they live in the compact closed category $\textbf{fHilb}$.
Each atomic pregroup type is associated a finite-dimensional Hilbert space, each individual (typed) word is associated a pure state in the Hilbert space associated to its type, and the pregroup grammatical structure is realised as an interaction between the word states mediated by certain entangling effects.

\subsection{Compositionality from Grammar}

\begin{definition}
    A \emph{pregroup} $\textbf{P}$ is the rigid category (with chosen duals) freely generated by a finite set $T$ of \emph{atomic types}. Specifically, the objects in $\textbf{P}$ are generated from $T$ as follows:
    \begin{itemize}
        \itemsep0mm
        \item every atomic type $\tau \in T$ is a type (aka object) in $\textbf{P}$;
        \item for every type $\tau$ in $\textbf{P}$, the \emph{left adjoint} $\tau^{l}$ and the \emph{right adjoint} $\tau^{r}$ are also types in $\textbf{P}$;
        \item for every pair of types $\tau, \sigma \in P$, the \emph{product type} $\tau \otimes \sigma$ is also a type $\textbf{P}$ (typically written $\tau\sigma$);
        \item the product operation is strictly associative and has a bilateral unit, the \emph{unit type} $\epsilon \in P$;
    \end{itemize}
    The pregroup $\textbf{P}$ is a poset category, i.e. every pair of types $A, B$ in $T$ as at most one morphism $A \rightarrow B$.
    As convention in poset categories, we write $A \leq B$ for the unique morphism $A \rightarrow B$, if it exists.
    The morphisms of $\textbf{P}$ are generated as follows:
    \begin{itemize}
        \itemsep0mm
        \item for every type $\tau$ in $\textbf{P}$, we have morphisms $\tau \tau^r \leq \epsilon$ and $\tau^l \tau \leq \epsilon$, known as \emph{contractions} or \emph{cups};
        \item for every type $\tau$ in $\textbf{P}$, we have morphisms $\epsilon \leq \tau^r \tau$ and $\epsilon \leq \tau \tau^l$, known as \emph{expansions} or \emph{caps};
    \end{itemize}
    All further equalities between objects and between morphisms follow from the requirement that $\textbf{P}$ be a poset category.
\end{definition}

\begin{remark}
    Some useful equalities which can be derived from the requirement that $\textbf{P}$ be a poset category include: the \emph{snake equations} between caps and caps; the cancellation of left and right duals, $(\tau^l)^r = \tau = (\tau^r)^l$; the stability of the unit type under duals, $\epsilon^l = \epsilon = \epsilon^r$; the interplay between duals and products, $(\tau \sigma)^l = \sigma^l \tau^l$ and $(\tau \sigma)^r = \sigma^r \tau^r$.
\end{remark}

We use a graphical calculus for autonomous categories to depict morphisms in a pregroup (which are also known as \emph{reductions}, following the tradition of rule-based grammar).
In particular, the contractions and expansions are depicted as follows:

\begin{equation}
\begin{tikzpicture}
	\begin{pgfonlayer}{nodelayer}
		\node [style=none] (0) at (-6, 0.5) {$\tau$};
		\node [style=none] (1) at (-4, 0.5) {$\tau^r$};
		\node [style=none] (2) at (-2, 0.5) {$\tau^l$};
		\node [style=none] (3) at (0, 0.5) {$\tau^{\phantom{0}}$};
		\node [style=none] (4) at (2, 0.5) {};
		\node [style=none] (5) at (4, 0.5) {};
		\node [style=none] (6) at (6, 0.5) {};
		\node [style=none] (7) at (8, 0.5) {};
		\node [style=none] (8) at (-6, 0) {};
		\node [style=none] (9) at (-4, 0) {};
		\node [style=none] (10) at (-2, 0) {};
		\node [style=none] (11) at (0, 0) {};
		\node [style=none] (12) at (2, 0) {$\tau^r$};
		\node [style=none] (13) at (4, 0) {$\tau^{\phantom{0}}$};
		\node [style=none] (14) at (6, 0) {$\tau^{\phantom{0}}$};
		\node [style=none] (15) at (8, 0) {$\tau^l$};
	\end{pgfonlayer}
	\begin{pgfonlayer}{edgelayer}
		\draw [bend right] (8.center) to (9.center);
		\draw [bend right] (10.center) to (11.center);
		\draw [bend left] (4.center) to (5.center);
		\draw [bend left] (6.center) to (7.center);
	\end{pgfonlayer}
\end{tikzpicture}
\end{equation}

\noindent
In our diagrams, objects are multiplied left-to-right and morphisms are composed top-to-bottom: in the above, the two morphism on the left are the caps/contractions and the two morphisms on the right are the cups/expansions.
The empty type is the tensor unit, hence it is not depicted.

\begin{definition}\label{def:PG}
    Given a pregroup $\textbf{P}$, a \emph{pregroup grammar} $\textbf{G}$ for $\textbf{P}$ is a pair $\textbf{G} = (L, t)$ consisting of a \emph{lexicon} $L$ (a finite set of \emph{words} $w \in L$) together with a \emph{typing map} $t: L \rightarrow \textbf{P}$ associating a pregroup type $t(w) \in P$ to each word $w \in L$.
\end{definition}

\noindent
When working with pregroup grammars, pregroup types subsume the role traditionally played by lexical categories such as nouns, adjectives, verbs, adverbs, etc~\cite{lambek2008word}.
If the lexicon is already POS-tagged by other means, then a pregroup grammar can be obtained by associating pregroup types to each POS-tag.
A pregroup grammar $\textbf{G}$ can equivalently be seen as the strict monoidal category generated from $\textbf{P}$ by adding states $\epsilon \rightarrow \tau$ labelled by the words $\suchthat{w \in L}{t(w) = \tau}$ for each individual type $\tau \in P$.
This is the same as saying that $\textbf{G}$ is the rigid category (with chosen duals) freely generated by the atomic types of $\textbf{P}$ and by states $w: \epsilon \rightarrow t(w)$ for all words $w \in L$.

\begin{remark}
    Taking the left/right duals gives monoidal functors $(\_)^l, (\_)^r: \textbf{P} \rightarrow \textbf{P}^{op}$ and by extension monoidal functors $(\_)^l, (\_)^r: \textbf{G} \rightarrow \textbf{G}^{op}$, where the opposite categories $\textbf{P}^{op}$ and $\textbf{G}^{op}$ (those with arrows reversed) are equipped with the opposite monoidal product $A \boxtimes B := B \otimes A$ with respect to the original categories $\textbf{P}$ and $\textbf{G}$.
\end{remark}

\begin{definition} \label{def:grammaticality}
    Consider a pregroup $\textbf{P}$ with atomic types $T$ is equipped with a chosen \emph{sentence type} $s \in T$. A \emph{grammatical sentence} is a non-empty sequence $\underline{w} = (w_1, ..., w_n)$ of words together with a sequence of contractions and expansions witnessing that the product type associated to the sequence reduces to the sentence type:

    \begin{equation}
    \prod_{i = 1}^n t(w_i) \leq s
    \end{equation}

    In this work, empty products are identified with the unit type $\epsilon$ and non-empty products are expanded left-to-right as $\prod_{i = 1}^n t(w_i) := t(w_1) \hdots t(w_n)$.
    \footnote{Such a choice of convention is made necessary by non-commutativity of the product operation on types.}
\end{definition}

Deciding \emph{grammaticality}---that is deciding whether the product type associated to a given sequence of words by a pregroup grammar $\textbf{G}$ reduces to the chosen sentence type $s$---is an efficiently solvable problem~\cite{preller,felice2019functorial}.
In the graphical calculus, the witness of grammaticality for a sequence of words is a pattern of nested caps (and possibly cups) connecting the atoms in the product type in such a way as to leave only one $s$ type open:

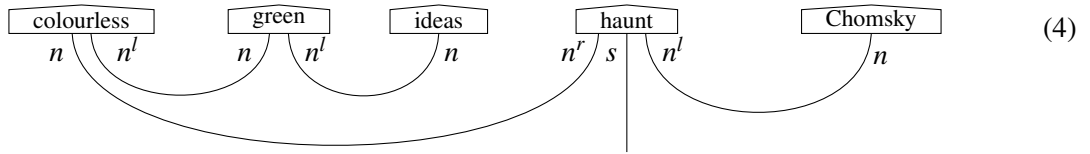
\begin{equation}
\begin{tikzpicture}
	\begin{pgfonlayer}{nodelayer}
		\node [style=none] (5) at (-11, 0) {};
		\node [style=none] (6) at (-10.5, 0) {};
		\node [style=none] (7) at (-5.75, 0) {};
		\node [style=none] (8) at (-5.25, 0) {};
		\node [style=none] (9) at (-1.25, 0) {};
		\node [style=none] (10) at (3.75, 0) {};
		\node [style=none] (11) at (3, 0) {};
		\node [style=none] (12) at (4.25, 0) {};
		\node [style=none] (13) at (10.25, 0) {};
		\node [style=none] (14) at (3.75, -3.25) {};
		\node [style=langstate] (15) at (-10.75, 0.25) {colourless};
		\node [style=langstate] (16) at (-5.5, 0.25) {green};
		\node [style=langstate] (17) at (-1.25, 0.25) {ideas};
		\node [style=langstate] (18) at (3.75, 0.25) {haunt};
		\node [style=langstate] (19) at (10.25, 0.25) {Chomsky};
		\node [style=none] (20) at (10.5, -0.75) {$n$};
		\node [style=none] (21) at (5, -0.5) {$n^l$};
		\node [style=none] (22) at (2.5, -0.5) {$n^r{}^{\phantom{0}}$};
		\node [style=none] (23) at (3.5, -0.5) {$s^{\phantom{0}}$};
		\node [style=none] (24) at (-0.75, -0.5) {$n^{\phantom{0}}$};
		\node [style=none] (25) at (-4.5, -0.5) {$n^l$};
		\node [style=none] (26) at (-6.25, -0.5) {$n^{\phantom{0}}$};
		\node [style=none] (27) at (-9.5, -0.5) {$n^l$};
		\node [style=none] (28) at (-11.25, -0.5) {$n^{\phantom{0}}$};
	\end{pgfonlayer}
	\begin{pgfonlayer}{edgelayer}
		\draw [bend right=90, looseness=1.25] (12.center) to (13.center);
		\draw [bend right=90, looseness=1.25] (6.center) to (7.center);
		\draw [bend right=90, looseness=1.50] (8.center) to (9.center);
		\draw [bend right=90, looseness=0.75] (5.center) to (11.center);
		\draw (10.center) to (14.center);
	\end{pgfonlayer}
\end{tikzpicture}\label{eq:grammar}
\end{equation}

\noindent
There exist several measures of how grammatical a sentence is, such as \emph{Harmony}.
For pregroup grammars, Harmony can be defined as the number of non-sentence atomic types left open after parsing~\cite{lewis2016harmonic}.
Harmony maximisation---i.e. finding the parsing closest to a witness of sentence grammaticality---is a non-trivial problem which can have a polynomial speed-up when it is approached with quantum annealing~\cite{wiebe2019quantum}.
Thus, in principle, the first part of the pipeline that prepares the input
diagram could be tackled with a quantum computer instead of
using classical methods.
However, this is left for future investigation
and we stay aligned with the DisCoCat literature
in which the pregroup parsing problem is assumed solved
and the sentence's diagram is given as input to the problem
of computing semantics.

\subsection{Distributional Meaning}

Given a pregroup grammar $\textbf{G}$, semantics for the grammar are given by monoidal functors $F: \textbf{G} \rightarrow \textbf{C}$, where $\textbf{C}$ is some suitable rigid category (with compact closed categories as a special case).
In \emph{distributional semantics}, dagger compact categories of finite-dimensional Hilbert spaces are often of interest, such as the category $\textbf{fHilb}$ of complex matrices used in this work or its real matrices analogue, used in most traditional approaches to NLP.
There is a vast literature on methods for associating distributional semantics to words, from the early bag-of-words approaches to more modern ones based on artificial neural networks~\cite{Milajevs2014}.
Non-vectorial representations have also appeared in the literature~\cite{Bolt2016,erkkatrin,vilnis2014word,Vilnis2018}.
Compositional distributional semantics has been successfully benchmarked against more traditional approaches, outperforming several of the contemporary techniques \cite{GrefSadr,KartSadr,KartsaklisSadrzadeh2014,wijnholds-sadrzadeh-2019-evaluating}.

We work in a presentation of $\textbf{fHilb}$ where objects are the positive integers---the possible dimensions for finite-dimensional Hilbert spaces---and morphisms $m \rightarrow n$ are $n$-by-$m$ complex matrices.
The presentation is made dagger compact by taking the conjugate transpose of matrices as the dagger, together with the following chosen duals:

\begin{enumerate}
    \item we pick our dual objects as $n^* := n$;
    \item we make a choice of orthonormal basis---the \emph{computational basis}---for all $n$ prime;
    \item we extend our chosen computational bases to all $n$ by considering product bases;
    \item we define the cap $n \otimes n \rightarrow 1$ by setting $\ket{e_i} \otimes \ket{e_j} \mapsto \delta_{ij}$ on the chosen orthonormal basis for $n$;
    \item we define the cup $1 \rightarrow n \otimes n$ as the adjoint of the cap $n \otimes n \rightarrow 1$.
\end{enumerate}

\noindent
With the above presentation, giving distributional semantics $\textbf{G} \rightarrow \textbf{fHilb}$ concretely means the following:

\begin{itemize}
    \itemsep0mm
    \item a finite dimension $d_\tau$ is associated to each atomic type $\tau \in T$ of the pregroup $\textbf{P}$;
    \footnote{Linearity and finite-dimensionality actually force the same dimension to be associated to all duals, i.e. $d_{\tau^l} = d_\tau = d_{\tau^r}$.}
    \item a $d_{t(w)}$-dimensional complex vector $\ket{w}$ is associated to each word $w \in L$.
\end{itemize}

\noindent
We refer to the data above as a \emph{word embedding}.
For example, the the word ``haunt'' would have a complex vector of dimension $d_{n^r} d_s d_{n^l}$ associated to it by a word embedding, which we can represent as follows in the graphical calculus:

\begin{equation}
    \begin{tikzpicture}
	\begin{pgfonlayer}{nodelayer}
		\node [style=langstate] (0) at (0, 2) {$\text{haunt}$};
		\node [style=none] (1) at (0, 0.5) {};
		\node [style=none] (2) at (-1, 0.5) {};
		\node [style=none] (3) at (1, 0.5) {};
		\node [style=none] (4) at (-1, 1.75) {};
		\node [style=none] (5) at (0, 1.75) {};
		\node [style=none] (6) at (1, 1.75) {};
		\node [style=none] (7) at (-1, -0.20) {$\small d_{n^r}$};
		\node [style=none] (9) at (0, -0.20) {\small $d_s$};
		\node [style=none] (10) at (1, -0.20) {\small $d_{n^l}$};
	\end{pgfonlayer}
	\begin{pgfonlayer}{edgelayer}
		\draw (5.center) to (1.center);
		\draw (4.center) to (2.center);
		\draw (6.center) to (3.center);
	\end{pgfonlayer}
\end{tikzpicture}
\end{equation}

\begin{remark}
    Because information about the factors of the dimension $d_{n^r} d_s d_{n^l}$ is derivable from the word embedding together with the word type $t(\text{haunt})$, we can equivalently treat the above as a vector in dimension $d_{n^r} d_s d_{n^l}$ or as a tensor of arity 3 in the individual dimensions $d_{n^r}$, $d_s$ and $d_{n^l}$.
    In general, the arity of the tensor associated to a word $w$ is the number of atomic types appearing in $t(w)$.
\end{remark}

\noindent
Given a word embedding, every pregroup grammatical parsing can be turned into tensor contraction by sending the caps and caps of $\textbf{G}$ to the chosen caps and caps of $\textbf{fHilb}$, as in the following example:

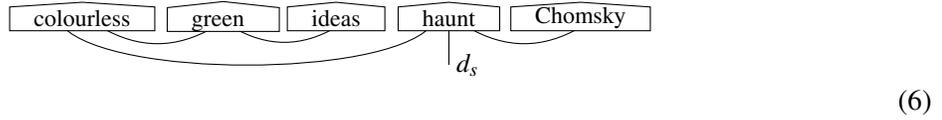
\begin{equation}\label{eq:chomsky-meaning}
\begin{tikzpicture}
	\begin{pgfonlayer}{nodelayer}
		\node [style=langstate] (0) at (3, 2.25) {haunt};
		\node [style=langstate] (1) at (6.5, 2.25) {Chomsky};
		\node [style=langstate] (2) at (0, 2.25) {ideas};
		\node [style=langstate] (3) at (-3, 2.25) {green$^{\phantom{1}}$};
		\node [style=langstate] (4) at (-6.75, 2.25) {colourless};
		\node [style=none] (5) at (-7.25, 2) {};
		\node [style=none] (6) at (-6.25, 2) {};
		\node [style=none] (7) at (-3.25, 2) {};
		\node [style=none] (8) at (-2.75, 2) {};
		\node [style=none] (9) at (0, 2) {};
		\node [style=none] (10) at (3, 2) {};
		\node [style=none] (11) at (2.5, 2) {};
		\node [style=none] (12) at (3.5, 2) {};
		\node [style=none] (13) at (6.5, 2) {};
		\node [style=none] (14) at (3, 1) {};
		\node [style=none] (15) at (3.5, 1) {};
		\node [style=none] (16) at (3.5, 1) {\small $d_s$};
	\end{pgfonlayer}
	\begin{pgfonlayer}{edgelayer}
		\draw [bend right] (12.center) to (13.center);
		\draw [bend right] (6.center) to (7.center);
		\draw [bend right] (8.center) to (9.center);
		\draw [bend right=45, looseness=0.50] (5.center) to (11.center);
		\draw (10.center) to (14.center);
	\end{pgfonlayer}
\end{tikzpicture}
\end{equation}

\noindent
As a special case, grammatical sentences find interpretation as $d_s$-dimensional vectors, as the witness of grammaticality results in contraction of all tensor legs except for one of type $s$.

In our presentation of $\textbf{fHilb}$, each Hilbert space $n$ has a chosen classical structure (i.e. a special commutative $\dagger$-Frobenius algebra) associated to it, corresponding to our choice of computational basis.
This means that spiders are available as additional ingredients to our semantics:

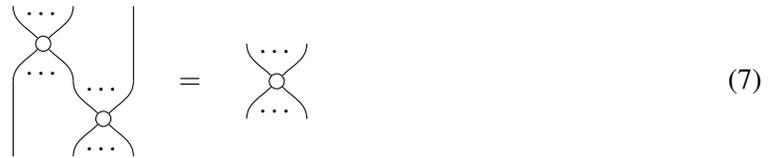
\begin{equation}
\begin{tikzpicture}[scale=0.80]
	\begin{pgfonlayer}{nodelayer}
		\node [style=none] (0) at (0, 0) {};
		\node [style=none] (1) at (-2, 0) {};
		\node [style=none] (2) at (-2, 2.5) {};
		\node [style=none] (3) at (0, 2.5) {};
		\node [style=none] (4) at (-1, 2.25) {\ldots};
		\node [style=none] (5) at (-1, 0.25) {\ldots};
		\node [style=white dot] (6) at (-1, 1.25) {};
		\node [style=none] (7) at (-1, 1.25) {};
		\node [style=none] (8) at (0, -2.5) {};
		\node [style=none] (9) at (1, -1.25) {};
		\node [style=none] (10) at (0, 0) {};
		\node [style=none] (11) at (1, -0.25) {\ldots};
		\node [style=none] (12) at (1, -2.25) {\ldots};
		\node [style=white dot] (13) at (1, -1.25) {};
		\node [style=none] (14) at (2, -2.5) {};
		\node [style=none] (15) at (2, 0) {};
		\node [style=none] (16) at (-2, -2.5) {};
		\node [style=none] (17) at (2, 2.5) {};
	\end{pgfonlayer}
	\begin{pgfonlayer}{edgelayer}
		\draw [style=swap, in=135, out=-90, looseness=1.00] (2.center) to (6);
		\draw [style=swap, in=45, out=-90, looseness=1.00] (3.center) to (6);
		\draw [style=swap, in=90, out=-45, looseness=1.00] (6) to (0.center);
		\draw [style=swap, in=90, out=-135, looseness=1.00] (6) to (1.center);
		\draw [style=swap, in=135, out=-90, looseness=1.00] (10.center) to (13);
		\draw [style=swap, in=45, out=-90, looseness=1.00] (15.center) to (13);
		\draw [style=swap, in=90, out=-45, looseness=1.00] (13) to (14.center);
		\draw [style=swap, in=90, out=-135, looseness=1.00] (13) to (8.center);
		\draw [style=swap] (17.center) to (15.center);
		\draw [style=swap] (1.center) to (16.center);
	\end{pgfonlayer}
\end{tikzpicture}
\hspace{5mm} = \hspace{5mm}
\begin{tikzpicture}[scale=0.80]
	\begin{pgfonlayer}{nodelayer}
		\node [style=none] (0) at (1, -1.25) {};
		\node [style=none] (1) at (-1, -1.25) {};
		\node [style=none] (2) at (-1, 1.25) {};
		\node [style=none] (3) at (1, 1.25) {};
		\node [style=none] (4) at (0, 1) {\ldots};
		\node [style=none] (5) at (0, -1) {\ldots};
		\node [style=white dot] (6) at (0, 0) {};
		\node [style=none] (7) at (0, 0) {};
	\end{pgfonlayer}
	\begin{pgfonlayer}{edgelayer}
		\draw [style=swap, in=135, out=-90, looseness=1.00] (2.center) to (6);
		\draw [style=swap, in=45, out=-90, looseness=1.00] (3.center) to (6);
		\draw [style=swap, in=90, out=-45, looseness=1.00] (6) to (0.center);
		\draw [style=swap, in=90, out=-135, looseness=1.00] (6) to (1.center); 
	\end{pgfonlayer}
\end{tikzpicture}
\end{equation}

\noindent
Spiders---with cups and caps as special two-legged cases---have been used in the past to associate semantics to functional and connective words---such as ``does", ``is" and ``are"---or to relative pronouns---such as ``which" and ``that" \cite{frobeniusanatomy,Sadrzadeh2013,Sadrzadeh2014}:

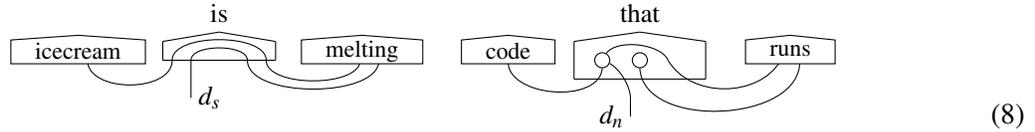
\begin{equation}\label{eq:capfrobansatz}
\begin{tikzpicture}
	\begin{pgfonlayer}{nodelayer}
		\node [style=langstate] (0) at (-3.75, 1.75) {icecream};
		\node [style=langstate] (1) at (3.8, 1.75) {melting};
		\node [style=none] (3) at (0, 2.75) {{\small is}};
		\node [style=none] (4) at (-3.5, 1.5) {};
		\node [style=none] (5) at (-1.25, 1.5) {};
		\node [style=none] (6) at (-0.75, 1.5) {};
		\node [style=none] (7) at (0.75, 1.5) {};
		\node [style=none] (8) at (1.25, 1.5) {};
		\node [style=none] (9) at (3.75, 1.5) {};
		\node [style=none] (10) at (4.25, 1.5) {};
		\node [style=none] (11) at (-0.75, 0.5) {};
		\node [style=none] (12) at (-1.5, 1.5) {};
		\node [style=none] (13) at (1.5, 1.5) {};
		\node [style=none] (14) at (1.5, 2) {};
		\node [style=none] (15) at (-1.5, 2) {};
		\node [style=none] (16) at (0, 2.25) {};
		\node [style=none] (17) at (-0.25, 0.5) {\small $d_s$};
	\end{pgfonlayer}
	\begin{pgfonlayer}{edgelayer}
		\draw [bend right=90] (4.center) to (5.center);
		\draw [bend left=90, looseness=0.75] (5.center) to (8.center);
		\draw [bend left=90, looseness=0.75] (6.center) to (7.center);
		\draw [bend right=90, looseness=0.75] (8.center) to (9.center);
		\draw [bend right=90, looseness=0.75] (7.center) to (10.center);
		\draw (6.center) to (11.center);
		\draw (15.center) to (16.center);
		\draw (16.center) to (14.center);
		\draw (14.center) to (13.center);
		\draw (13.center) to (12.center);
		\draw (12.center) to (15.center);
	\end{pgfonlayer}
\end{tikzpicture}
\hspace{5mm}
\begin{tikzpicture}
	\begin{pgfonlayer}{nodelayer}
		\node [style=langstate] (0) at (-3.5, 1.75) {code};
		\node [style=langstate] (1) at (4, 1.75) {runs};
		\node [style=none] (3) at (0, 2.75) {{\small that}};
		\node [style=none] (4) at (-3.5, 1.5) {};
		\node [style=none] (9) at (3.75, 1.5) {};
		\node [style=none] (10) at (4.25, 1.5) {};
		\node [style=white dot] (11) at (-1, 1.5) {};
		\node [style=white dot] (12) at (0, 1.5) {};
		\node [style=none] (13) at (0.75, 1.5) {};
		\node [style=none] (14) at (-0.25, 0.5) {};
		\node [style=none] (15) at (-0.25, 0) {};
		\node [style=none] (16) at (-1.75, 1) {};
		\node [style=none] (17) at (1.75, 1) {};
		\node [style=none] (18) at (-1.75, 2) {};
		\node [style=none] (19) at (0, 2.25) {};
		\node [style=none] (20) at (1.75, 2) {};
		\node [style=none] (21) at (-1, 0) {};
		\node [style=none] (22) at (-.75, 0) {\small $d_n$};
	\end{pgfonlayer}
	\begin{pgfonlayer}{edgelayer}
		\draw [bend left=60] (9.center) to (13.center);
		\draw [bend right=90] (12) to (10.center);
		\draw [bend left=90] (11) to (4.center);
		\draw [bend left, looseness=0.75] (11) to (14.center);
		\draw [bend left=60, looseness=0.75] (11) to (13.center);
		\draw (15.center) to (14.center);
		\draw (16.center) to (17.center);
		\draw (20.center) to (17.center);
		\draw (19.center) to (20.center);
		\draw (18.center) to (19.center);
		\draw (16.center) to (18.center);
	\end{pgfonlayer}
\end{tikzpicture}
\end{equation}


\section{Diagram Rewriting}

On the way to quantum circuits, we need to simplify our diagrams to optimise our ultimate use of quantum resources.
Specifically, we present two diagram simplification methods which aim to reduce circuit width and depth independently of the choice of word ans\"atze.
Both methods require additional flexibility in the manipulation of diagrams and hence take place in the following \emph{symmetric} version of the pregroup grammar.

\begin{definition}
    A \emph{symmetric pregroup grammar} $\hat{\textbf{G}}$ is the compact closed category obtained by introducing symmetry isomorphisms to a pregroup grammar $\textbf{G}$, keeping the same objects.
\end{definition}

\noindent
If $\textbf{G}$ is a pregroup grammar and $\hat{\textbf{G}}$ is the associated symmetric pregroup grammar, then there is a faithful monoidal functor $i_{\textbf{G}}: \textbf{G} \rightarrow \hat{\textbf{G}}$ which is the identity on objects.
Any monoidal functor $F: \textbf{G} \rightarrow \textbf{C}$ towards a compact closed category $\textbf{C}$ factors as $F = \hat{F} \circ i_{\textbf{G}}$ for a unique monoidal functor $\hat{F}: \hat{\textbf{G}} \rightarrow \textbf{C}$.
As a consequence of this observation, the introduction of a symmetric pregroup grammar provides additional degrees of freedom when it comes to diagram rewriting, without imposing any additional restrictions to the compact closed semantics traditionally considered in the DisCoCat framework.

\subsection{The \texttt{bigraph} method}

The first rewrite method we present, which we call the \texttt{bigraph} method, completes and improves the original Zeng-Coecke algorithm \cite{Zeng2016}.
We start with the simplest scenario, described in the original algorithm: the diagram has a single open wire (e.g. it is a grammatical sentence) and the cups/caps connect words in such a way as to form a an acyclic (undirected) graph.
For example, we could consider the grammatical sentence from \eqref{eq:grammar}.

As its first step, the \texttt{bigraph} method turns the diagram into a bipartite graph, based on the distance from the ``root'' word, which is defined to be the one connected to the unique open wire.
Words at even distance from the root are left in place as states, while words at odd distance from the root are transposed into effects:

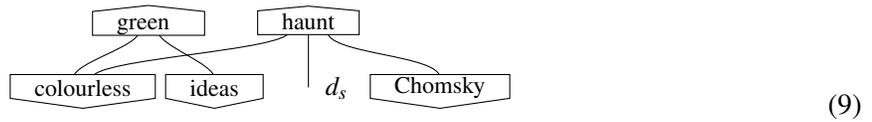
\begin{equation}\label{eq:chomskyBigraph}
\begin{tikzpicture}
	\begin{pgfonlayer}{nodelayer}
		\node [style=langstate] (0) at (3, 2.25) {haunt};
		\node [style=langeffect] (1) at (6.5, 0.5) {Chomsky};
		\node [style=langeffect] (2) at (0.5, 0.5) {ideas};
		\node [style=langstate] (3) at (-1.25, 2.25) {green$^{\phantom{1}}$};
		\node [style=langeffect] (4) at (-3, 0.5) {colourless};
		\node [style=none] (5) at (-2.75, 0.75) {};
		\node [style=none] (6) at (-3.25, 0.75) {};
		\node [style=none] (7) at (-1.5, 2) {};
		\node [style=none] (8) at (-1, 2) {};
		\node [style=none] (9) at (0.5, 0.75) {};
		\node [style=none] (10) at (3, 2) {};
		\node [style=none] (11) at (2.5, 2) {};
		\node [style=none] (12) at (3.5, 2) {};
		\node [style=none] (13) at (6.5, 0.75) {};
		\node [style=none] (14) at (3, 0.5) {};
		\node [style=none] (16) at (3.75, 0.5) {\small $d_s$};
	\end{pgfonlayer}
	\begin{pgfonlayer}{edgelayer}
		\draw [in=90, out=-90, looseness=0.50] (12.center) to (13.center);
		\draw [in=270, out=90, looseness=0.50] (6.center) to (7.center);
		\draw [in=90, out=-90, looseness=0.50] (8.center) to (9.center);
		\draw [in=-90, out=90, looseness=0.25] (5.center) to (11.center);
		\draw (10.center) to (14.center);
	\end{pgfonlayer}
\end{tikzpicture}
\end{equation}

\noindent
This is essentially the method originally described in \cite{Zeng2016}, except that the transpose in the computational basis is used to turn states into effects, instead of the dagger used in the original formulation:

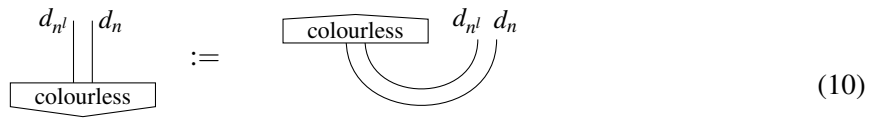
\begin{equation}\label{eq:colourlessTranspose}
\begin{tikzpicture}
	\begin{pgfonlayer}{nodelayer}
		\node [style=langeffect] (0) at (-2, -0.25) {colourless};
		\node [style=none] (1) at (-2.25, 0) {};
		\node [style=none] (2) at (-1.75, 0) {};
		\node [style=none] (3) at (-2.25, 1.75) {};
		\node [style=none] (4) at (-1.75, 1.75) {};
		\node [style=none] (5) at (-2.75, 1.75) {\small $d_{n^l}$};
		\node [style=none] (7) at (-1.25, 1.75) {\small $d_n$};
		\node [style=none] (8) at (1.25, 0.75) {$:=$};
		\node [style=none] (9) at (8.5, 1.25) {};
		\node [style=none] (10) at (9, 1.25) {};
		\node [style=langstate] (11) at (5.25, 1.5) {colourless};
		\node [style=none] (12) at (5, 1.25) {};
		\node [style=none] (13) at (5.5, 1.25) {};
		\node [style=none] (14) at (8.25, 1.75) {\small $d_{n^l}$};
		\node [style=none] (15) at (9.25, 1.75) {\small $d_n$};
	\end{pgfonlayer}
	\begin{pgfonlayer}{edgelayer}
		\draw (1.center) to (3.center);
		\draw (4.center) to (2.center);
		\draw [bend right=90, looseness=1.50] (13.center) to (9.center);
		\draw [bend right=90, looseness=1.50] (12.center) to (10.center);
	\end{pgfonlayer}
\end{tikzpicture}
\end{equation}

\noindent
One issue not originally foreseen with this approach is the introduction of wire crossings.
This is a problem when it comes to implementation on NISQ devices: a swap between neighbouring qubits involves up to three entangling gates, which in turn lead to significant increase in circuit depths and exponential decrease in fidelity.

To tackle this issue, the \texttt{bigraph} method attempts to minimise the number of crossings by altering the linear order of words in the two classes.
\footnote{Because the semantic relationships between words are now encoded in the tensor contractions, their linear ordering is no longer relevant and can be used as an additional degree of freedom in the optimisation.}
For example, consider the following grammatical sentence:

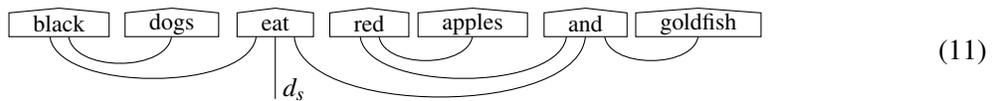
\begin{equation}\label{eq:badDog}
\begin{tikzpicture}
	\begin{pgfonlayer}{nodelayer}
		\node [style=langstate] (0) at (-3.25, 0.75) {black};
		\node [style=langstate] (1) at (-0.25, 0.75) {dogs};
		\node [style=langstate] (3) at (2.5, 0.75) {eat};
		\node [style=langstate] (4) at (5, 0.75) {red};
		\node [style=langstate] (9) at (13.75, 0.75) {goldfish};
		\node [style=langstate] (10) at (10.75, 0.75) {and};
		\node [style=langstate] (12) at (7.75, 0.75) {apples};
		\node [style=none] (13) at (-3.5, 0.5) {};
		\node [style=none] (14) at (-0.25, 0.5) {};
		\node [style=none] (16) at (2.5, 0.5) {};
		\node [style=none] (17) at (4.75, 0.5) {};
		\node [style=none] (18) at (7.75, 0.5) {};
		\node [style=none] (20) at (10.75, 0.5) {};
		\node [style=none] (21) at (13.75, 0.5) {};
		\node [style=none] (22) at (-3, 0.5) {};
		\node [style=none] (26) at (3, 0.5) {};
		\node [style=none] (27) at (5.25, 0.5) {};
		\node [style=none] (29) at (10.25, 0.5) {};
		\node [style=none] (30) at (11.25, 0.5) {};
		\node [style=none] (31) at (2, 0.5) {};
		\node [style=none] (32) at (2.5, -1.25) {};
		\node [style=none] (33) at (3, -1) {\small $d_s$};
	\end{pgfonlayer}
	\begin{pgfonlayer}{edgelayer}
		\draw [bend right=90] (22.center) to (14.center);
		\draw [bend right=90] (27.center) to (18.center);
		\draw [bend right=90] (30.center) to (21.center);
		\draw [bend right=90, looseness=0.75] (17.center) to (29.center);
		\draw [bend left=90, looseness=0.75] (31.center) to (13.center);
		\draw (16.center) to (32.center);
		\draw [bend right=90, looseness=0.75] (26.center) to (20.center);
	\end{pgfonlayer}
\end{tikzpicture}
\end{equation}

\noindent
After the initial transposition step, the following bipartite graph drawing is obtained:

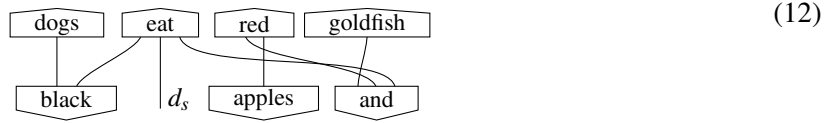
\begin{equation}\label{eq:badDogTranspose}
\begin{tikzpicture}
	\begin{pgfonlayer}{nodelayer}
		\node [style=langeffect] (0) at (0, -2.25) {black};
		\node [style=langstate] (1) at (-0.25, -0.25) {dogs};
		\node [style=langstate] (3) at (2.5, -0.25) {eat};
		\node [style=langstate] (4) at (5, -0.25) {red};
		\node [style=langstate] (9) at (8, -0.25) {goldfish};
		\node [style=langeffect] (10) at (8.25, -2.25) {and};
		\node [style=langeffect] (12) at (5.25, -2.25) {apples};
		\node [style=none] (13) at (0.25, -2) {};
		\node [style=none] (14) at (-0.25, -0.5) {};
		\node [style=none] (16) at (2.5, -0.5) {};
		\node [style=none] (17) at (4.75, -0.5) {};
		\node [style=none] (18) at (5.25, -2) {};
		\node [style=none] (20) at (8.75, -2) {};
		\node [style=none] (21) at (8, -0.5) {};
		\node [style=none] (22) at (-0.25, -2) {};
		\node [style=none] (26) at (3, -0.5) {};
		\node [style=none] (27) at (5.25, -0.5) {};
		\node [style=none] (29) at (8.25, -2) {};
		\node [style=none] (30) at (7.75, -2) {};
		\node [style=none] (31) at (2, -0.5) {};
		\node [style=none] (32) at (2.5, -2.5) {};
		\node [style=none] (33) at (3, -2.25) {\small $d_s$};
	\end{pgfonlayer}
	\begin{pgfonlayer}{edgelayer}
		\draw [in=-90, out=90, looseness=0.50] (22.center) to (14.center);
		\draw [in=90, out=-90, looseness=0.50] (27.center) to (18.center);
		\draw [in=-90, out=90, looseness=0.75] (30.center) to (21.center);
		\draw [in=90, out=-90, looseness=0.50] (17.center) to (29.center);
		\draw [in=90, out=-90, looseness=0.50] (31.center) to (13.center);
		\draw (16.center) to (32.center);
		\draw [in=90, out=-90, looseness=0.50] (26.center) to (20.center);
	\end{pgfonlayer}
\end{tikzpicture}
\end{equation}

\noindent
The drawing above involves 5 crossings: if each wire is mapped to a qubit and we use a reasonably optimised implementation of swaps between non-adjacent qubits, the crossings alone would increase the circuit depth by about 8 CNOTs.
However, a simple re-ordering of the words in the two classes leads to a graph drawing involving a single crossing:

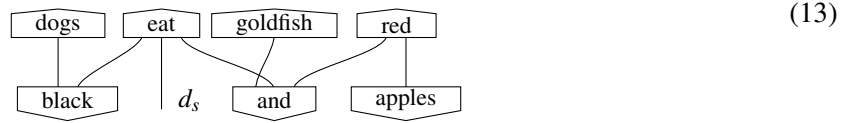
\begin{equation}\label{eq:badDogTransposeRearranged}
\begin{tikzpicture}
	\begin{pgfonlayer}{nodelayer}
		\node [style=langeffect] (0) at (0, -2.25) {black};
		\node [style=langstate] (1) at (-0.25, -0.25) {dogs};
		\node [style=langstate] (3) at (2.5, -0.25) {eat};
		\node [style=langstate] (4) at (8.75, -0.25) {red};
		\node [style=langstate] (9) at (5.5, -0.25) {goldfish};
		\node [style=langeffect] (10) at (5.5, -2.25) {and};
		\node [style=langeffect] (12) at (9, -2.25) {apples};
		\node [style=none] (13) at (0.25, -2) {};
		\node [style=none] (14) at (-0.25, -0.5) {};
		\node [style=none] (16) at (2.5, -0.5) {};
		\node [style=none] (17) at (8.5, -0.5) {};
		\node [style=none] (18) at (9, -2) {};
		\node [style=none] (20) at (6, -2) {};
		\node [style=none] (21) at (5.5, -0.5) {};
		\node [style=none] (22) at (-0.25, -2) {};
		\node [style=none] (26) at (3, -0.5) {};
		\node [style=none] (27) at (9, -0.5) {};
		\node [style=none] (29) at (5.5, -2) {};
		\node [style=none] (30) at (5, -2) {};
		\node [style=none] (31) at (2, -0.5) {};
		\node [style=none] (32) at (2.5, -2.5) {};
		\node [style=none] (33) at (3.25, -2.25) {\small $d_s$};
	\end{pgfonlayer}
	\begin{pgfonlayer}{edgelayer}
		\draw [in=-90, out=90, looseness=0.50] (22.center) to (14.center);
		\draw [in=90, out=-90, looseness=0.50] (27.center) to (18.center);
		\draw [in=-90, out=90, looseness=0.75] (30.center) to (21.center);
		\draw [in=90, out=-90, looseness=0.50] (17.center) to (20.center);
		\draw [in=90, out=-90, looseness=0.50] (31.center) to (13.center);
		\draw (16.center) to (32.center);
		\draw [in=90, out=-90, looseness=0.50] (26.center) to (29.center);
	\end{pgfonlayer}
\end{tikzpicture}
\end{equation}

\noindent
The \texttt{bigraph} method does not prescribe a specific algorithm to use when minimizing crossings: this is because the general problem of minimizing crossings in the planar drawing of bipartite graphs is NP-complete~\cite{Garey1983CrossingNI}.
\footnote{It is an open question whether the bipartite graphs that arise from pregroup grammars form a sub-class which is sufficiently restricted---e.g. due to the localised range of the connections---to bring the complexity down to P.}

The \texttt{bigraph} method relies on ans\"atze which can be easily transposed: failing that, each transposition would naively involve a doubling of the number of qubits and the preparation or measurement of nested bell states.
In Section~\ref{section:diagram2circuit} we shall see that our chosen ans\"atze have this property.
In fact, circuit ans\"atze such as those used in this work often have more symmetries, which can be used to further optimise the resulting quantum circuit.
For example, it is easy to transform them in such a way as to reverse the order of their outputs: this means that any combination of swaps resulting in a complete reversal of the outputs of a single word is not going to ultimately increase the depth of the quantum circuit.
For example, the optimal arrangement for \eqref{eq:chomskyBigraph} using this additional assumption is as follows:

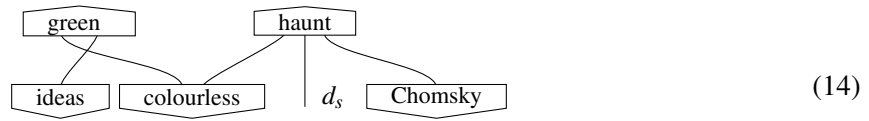
\begin{equation}\label{eq:chomskyBigraphOpt}
\begin{tikzpicture}
	\begin{pgfonlayer}{nodelayer}
		\node [style=langstate] (0) at (3, 1.75) {haunt};
		\node [style=langeffect] (1) at (6.5, -0.25) {Chomsky};
		\node [style=langeffect] (2) at (-3.5, -0.25) {ideas};
		\node [style=langstate] (3) at (-3, 1.75) {green$^{\phantom{1}}$};
		\node [style=langeffect] (4) at (0, -0.25) {colourless};
		\node [style=none] (5) at (0.25, 0) {};
		\node [style=none] (6) at (-0.25, 0) {};
		\node [style=none] (7) at (-3.5, 1.5) {};
		\node [style=none] (8) at (-2.5, 1.5) {};
		\node [style=none] (9) at (-3.5, 0) {};
		\node [style=none] (10) at (3, 1.5) {};
		\node [style=none] (11) at (2.5, 1.5) {};
		\node [style=none] (12) at (3.5, 1.5) {};
		\node [style=none] (13) at (6.5, 0) {};
		\node [style=none] (14) at (3, -0.5) {};
		\node [style=none] (16) at (3.75, -0.25) {\small $d_s$};
	\end{pgfonlayer}
	\begin{pgfonlayer}{edgelayer}
		\draw [in=90, out=-90, looseness=0.50] (12.center) to (13.center);
		\draw [in=270, out=90, looseness=0.50] (6.center) to (7.center);
		\draw [in=90, out=-90, looseness=0.50] (8.center) to (9.center);
		\draw [in=-90, out=90, looseness=0.25] (5.center) to (11.center);
		\draw (10.center) to (14.center);
	\end{pgfonlayer}
\end{tikzpicture}
\end{equation}

In a more general scenario, a pregroup grammatical parsing might: (i) involve cups/expansions as well as caps/contractions; (ii) result in a cyclic graph.
The presence of cups/expansions is a non-issue when it comes to semantics in compact closed symmetric monoidal categories: thanks to the existence of symmetry isomorphisms, all cup will be cancelled out by caps in the target category.
The case of cyclic graphs requires more careful handling.
For example, consider the following odd parsing:

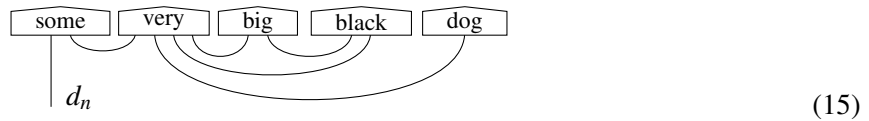
\begin{equation}\label{eq:cyclicParsing}
\begin{tikzpicture}
	\begin{pgfonlayer}{nodelayer}
		\node [style=langstate] (0) at (-4.5, 2.25) {some};
		\node [style=none] (1) at (-4.25, 2) {};
		\node [style=none] (2) at (-4.75, 2) {};
		\node [style=langstate] (4) at (-1.75, 2.25) {very};
		\node [style=none] (5) at (-2, 2) {};
		\node [style=none] (6) at (-2.5, 2) {};
		\node [style=none] (7) at (-1.5, 2) {};
		\node [style=none] (8) at (-1, 2) {};
		\node [style=langstate] (9) at (0.75, 2.25) {big};
		\node [style=none] (10) at (0.5, 2) {};
		\node [style=none] (12) at (1, 2) {};
		\node [style=langstate] (13) at (3.5, 2.25) {black};
		\node [style=none] (14) at (3.25, 2) {};
		\node [style=none] (15) at (3.75, 2) {};
		\node [style=langstate] (16) at (6.25, 2.25) {dog};
		\node [style=none] (17) at (6.25, 2) {};
		\node [style=none] (18) at (-4.75, 0) {};
		\node [style=none] (19) at (-4, 0.25) {$d_n$};
	\end{pgfonlayer}
	\begin{pgfonlayer}{edgelayer}
		\draw (18.center) to (2.center);
		\draw [bend right=90] (1.center) to (6.center);
		\draw [bend right=90, looseness=0.75] (5.center) to (17.center);
		\draw [bend left=90, looseness=0.75] (15.center) to (7.center);
		\draw [bend right=90, looseness=1.50] (8.center) to (10.center);
		\draw [bend right=90] (12.center) to (14.center);
	\end{pgfonlayer}
\end{tikzpicture}
\end{equation}

\noindent
In the presence of cycles, distance from the root is no longer a well-defined notion and it may not be possible to re-arrange the diagram as to form a bipartite graph.
Given any partition of words into two linearly ordered classes---a ``pseudo-bipartite'' drawing, so to speak---each edge between words of same class can be ``dragged'' to the other side as if it were a word, increasing the circuit width by two wires.
In other words, states can be turned into effects by transposing a state into an effect and vice versa; transposition graphically amounts to performing an 180 degree \emph{rotation}.
Such transpositions introduce wire crossings, of course.
This can be seen in the following re-arrangement for the odd parsing above:

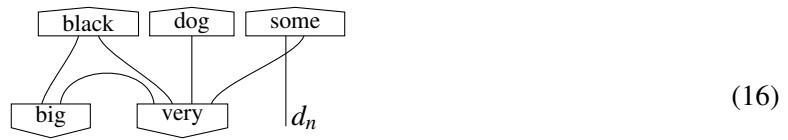
\begin{equation}\label{eq:cyclicParsingBigraph}
\begin{tikzpicture}
	\begin{pgfonlayer}{nodelayer}
		\node [style=langstate] (0) at (2, 2) {some};
		\node [style=none] (1) at (2.25, 1.75) {};
		\node [style=none] (2) at (1.75, 1.75) {};
		\node [style=langeffect] (4) at (-1, -0.5) {very};
		\node [style=none] (5) at (-0.75, -0.25) {};
		\node [style=none] (6) at (-0.25, -0.25) {};
		\node [style=none] (7) at (-1.25, -0.25) {};
		\node [style=none] (8) at (-1.75, -0.25) {};
		\node [style=langeffect] (9) at (-4.5, -0.5) {big};
		\node [style=none] (10) at (-4.25, -0.25) {};
		\node [style=none] (12) at (-4.75, -0.25) {};
		\node [style=langstate] (13) at (-3.5, 2) {black};
		\node [style=none] (14) at (-3.75, 1.75) {};
		\node [style=none] (15) at (-3.25, 1.75) {};
		\node [style=langstate] (16) at (-0.75, 2) {dog};
		\node [style=none] (17) at (-0.75, 1.75) {};
		\node [style=none] (18) at (1.75, -0.75) {};
		\node [style=none] (19) at (2.25, -0.5) {$d_n$};
	\end{pgfonlayer}
	\begin{pgfonlayer}{edgelayer}
		\draw (18.center) to (2.center);
		\draw [in=90, out=-90, looseness=0.50] (1.center) to (6.center);
		\draw [in=-90, out=90, looseness=0.50] (5.center) to (17.center);
		\draw [in=90, out=-90, looseness=0.50] (15.center) to (7.center);
		\draw [bend right=90, looseness=1.25] (8.center) to (10.center);
		\draw [in=-90, out=90, looseness=0.50] (12.center) to (14.center);
	\end{pgfonlayer}
\end{tikzpicture}
\end{equation}

\noindent
In this more general scenario, a cost function is required by the \texttt{bigraph} method to establish a trade-off between minimising the number of crossings and minimising the number of intra-class edges in a ``pseudo-bipartite'' drawing of the diagram.

Having handled all of the above, there is a final issue to consider: when dealing with parsings other than grammatical sentences, it is not necessary (nor necessarily desirable) that a single wire be left open.
To handle this most general scenario, the \texttt{bigraph} method operates as in the cyclic case---i.e. looks for a bipartite drawing optimising some trade-off between lack of crossings and lack of intra-class edges---but restricting the partitions in such a way that all words having one or more open wires are placed in the same class.
This ensures that the result always be a state, as was the case so far.\\

\noindent
(A Python implementation of the \texttt{bigraph} method will be available at \href{https://github.com/hashberg-io/qnlp}{github.com/hashberg-io/qnlp}.)

\subsection{The \texttt{snake_removal} method}

The second rewrite method we present, which we call the \texttt{snake_removal} method, is based on previous results by Ref.~\cite{dunnvicary}~and~\cite{Delpeuch2017}.
Instead of working with the full symmetric pregroup grammar $\hat{\textbf{G}}$, the \texttt{snake_removal} method considers the full sub-category of $\hat{\textbf{G}}$ spanned only by the atomic types and their products, which we call $\hat{\textbf{G}}_{aut}$.
This subcategory does not contain any word states which involve any adjoint types: instead, it contains the partial transposes of those states where all output wires with adjoint type have been bent into input wires.
A set of generators for this sub-category can be obtained by picking one representative for each word.
This procedure is called the \emph{autonomisation} of diagrams~\cite{Delpeuch2017}.
and examples of its application to the parts of speech of noun, adjective, intransitive verb, transitive verb, and relative pronoun are:
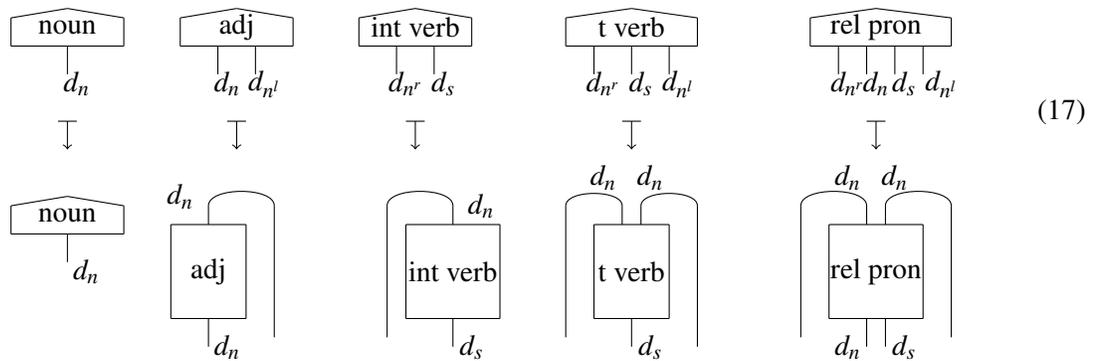
\begin{equation}
\input{./figures/tikz/autonomisation1.tikz}
\end{equation}
The idea is that states which have a wire of (left or right)
adjoint type are replaced by processes all of whose wires have non-adjoint types.
We understand this is possible simply by bending wires:
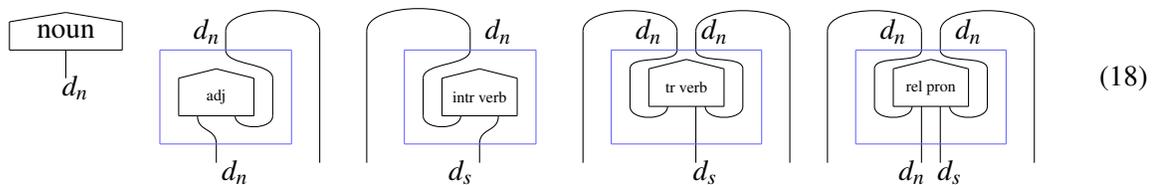
\begin{equation}
\begin{tikzpicture}
	\begin{pgfonlayer}{nodelayer}
		\node [style=none] (139) at (-11.5, 0.75) {};
		\node [style=none] (140) at (-8.5, 0.75) {};
		\node [style=none] (141) at (-8.5, 1.5) {};
		\node [style=none] (142) at (-11.5, 1.5) {};
		\node [style=none] (143) at (-10, 1.75) {};
		\node [style=none] (144) at (-10, 1.25) {noun};
		\node [style=none] (145) at (-10, 0.75) {};
		\node [style=none] (146) at (-10, 0) {};
		\node [style=none] (147) at (-9.75, -0.25) {$d_n$};
		\node [style=none] (148) at (-7, 0) {};
		\node [style=none] (149) at (-7, -1) {};
		\node [style=none] (150) at (-5, -1) {};
		\node [style=none] (151) at (-5, 0) {};
		\node [style=none] (152) at (-6, 0.25) {};
		\node [style=none] (153) at (-6, -0.5) {\tiny adj};
		\node [style=none] (154) at (-6.5, -1) {};
		\node [style=none] (155) at (-5.5, -1) {};
		\node [style=none] (158) at (-5.5, -2.5) {$d_n$};
		\node [style=none] (159) at (-6.25, 1.25) {$d_{n}$};
		\node [style=none] (160) at (-4.5, -1) {};
		\node [style=none] (161) at (-4.5, 0) {};
		\node [style=none] (162) at (-5.75, 0.75) {};
		\node [style=none] (163) at (-6, -1.75) {};
		\node [style=none] (164) at (-4, -1.75) {};
		\node [style=none] (165) at (-7.5, -1.75) {};
		\node [style=none] (166) at (-7.5, 0.75) {};
		\node [style=none] (167) at (-4, 0.75) {};
		\node [style=none] (168) at (-5.75, 1.25) {};
		\node [style=none] (169) at (-6, -2.25) {};
		\node [style=none] (170) at (2, 0) {};
		\node [style=none] (171) at (2, -1) {};
		\node [style=none] (172) at (0, -1) {};
		\node [style=none] (173) at (0, 0) {};
		\node [style=none] (174) at (1, 0.25) {};
		\node [style=none] (175) at (1, -0.5) {\tiny intr verb};
		\node [style=none] (176) at (1.5, -1) {};
		\node [style=none] (177) at (0.5, -1) {};
		\node [style=none] (178) at (0.5, -2.5) {$d_s$};
		\node [style=none] (179) at (1.5, 1.25) {$d_{n}$};
		\node [style=none] (180) at (-0.5, -1) {};
		\node [style=none] (181) at (-0.5, 0) {};
		\node [style=none] (182) at (0.75, 0.75) {};
		\node [style=none] (183) at (1, -1.75) {};
		\node [style=none] (184) at (-1, -1.75) {};
		\node [style=none] (185) at (2.5, -1.75) {};
		\node [style=none] (186) at (2.5, 0.75) {};
		\node [style=none] (187) at (-1, 0.75) {};
		\node [style=none] (188) at (0.75, 1.25) {};
		\node [style=none] (189) at (1, -2.25) {};
		\node [style=none] (192) at (7.5, 0.25) {};
		\node [style=none] (193) at (7.5, -0.75) {};
		\node [style=none] (194) at (5.5, -0.75) {};
		\node [style=none] (195) at (5.5, 0.25) {};
		\node [style=none] (196) at (6.5, 0.5) {};
		\node [style=none] (197) at (6.5, -0.25) {\tiny tr verb};
		\node [style=none] (198) at (6.75, -0.75) {};
		\node [style=none] (199) at (6, -0.75) {};
		\node [style=none] (200) at (7, -2.5) {$d_s$};
		\node [style=none] (201) at (5.5, 1.25) {$d_{n}$};
		\node [style=none] (202) at (5, -0.75) {};
		\node [style=none] (203) at (5, 0.25) {};
		\node [style=none] (204) at (6.25, 0.75) {};
		\node [style=none] (206) at (4.5, -1.75) {};
		\node [style=none] (207) at (8.5, -1.75) {};
		\node [style=none] (208) at (8.5, 0.75) {};
		\node [style=none] (209) at (4.5, 0.75) {};
		\node [style=none] (210) at (6.25, 1.25) {};
		\node [style=none] (213) at (7, -0.75) {};
		\node [style=none] (214) at (8, -0.75) {};
		\node [style=none] (215) at (8, 0.25) {};
		\node [style=none] (216) at (6.75, 0.75) {};
		\node [style=none] (217) at (6.75, 1.25) {};
		\node [style=none] (218) at (6.75, -2.25) {};
		\node [style=none] (219) at (14, 0.25) {};
		\node [style=none] (220) at (14, -0.75) {};
		\node [style=none] (221) at (12, -0.75) {};
		\node [style=none] (222) at (12, 0.25) {};
		\node [style=none] (223) at (13, 0.5) {};
		\node [style=none] (224) at (13, -0.25) {\tiny rel pron};
		\node [style=none] (225) at (13.25, -0.75) {};
		\node [style=none] (226) at (12.5, -0.75) {};
		\node [style=none] (227) at (13.5, -2.5) {$d_s$};
		\node [style=none] (228) at (12, 1.25) {$d_{n}$};
		\node [style=none] (229) at (11.5, -0.75) {};
		\node [style=none] (230) at (11.5, 0.25) {};
		\node [style=none] (231) at (12.75, 0.75) {};
		\node [style=none] (232) at (11, -1.75) {};
		\node [style=none] (233) at (15, -1.75) {};
		\node [style=none] (234) at (15, 0.75) {};
		\node [style=none] (235) at (11, 0.75) {};
		\node [style=none] (236) at (12.75, 1.25) {};
		\node [style=none] (238) at (13.5, -0.75) {};
		\node [style=none] (239) at (14.5, -0.75) {};
		\node [style=none] (240) at (14.5, 0.25) {};
		\node [style=none] (241) at (13.25, 0.75) {};
		\node [style=none] (242) at (13.25, 1.25) {};
		\node [style=none] (243) at (13.25, -2.25) {};
		\node [style=none] (244) at (12.75, -0.75) {};
		\node [style=none] (245) at (12.75, -2.25) {};
		\node [style=none] (246) at (7.25, 1.25) {$d_n$};
		\node [style=none] (247) at (14, 1.25) {$d_n$};
		\node [style=none] (248) at (12.5, -2.5) {$d_n$};
		\node [style=none] (249) at (-3.25, 1.25) {};
		\node [style=none] (250) at (-3.25, -2.25) {};
		\node [style=none] (251) at (-2, 1.25) {};
		\node [style=none] (252) at (-2, -2.25) {};
		\node [style=none] (253) at (3.75, 1.25) {};
		\node [style=none] (254) at (3.75, -2.25) {};
		\node [style=none] (255) at (9.25, 1.25) {};
		\node [style=none] (256) at (9.25, -2.25) {};
		\node [style=none] (257) at (10.25, 1.25) {};
		\node [style=none] (258) at (10.25, -2.25) {};
		\node [style=none] (259) at (15.75, 1.25) {};
		\node [style=none] (260) at (15.75, -2.25) {};
	\end{pgfonlayer}
	\begin{pgfonlayer}{edgelayer}
		\draw (139.center) to (145.center);
		\draw (145.center) to (140.center);
		\draw (140.center) to (141.center);
		\draw (143.center) to (141.center);
		\draw (143.center) to (142.center);
		\draw (142.center) to (139.center);
		\draw (145.center) to (146.center);
		\draw (149.center) to (154.center);
		\draw (154.center) to (155.center);
		\draw (155.center) to (150.center);
		\draw (150.center) to (151.center);
		\draw (151.center) to (152.center);
		\draw (152.center) to (148.center);
		\draw (148.center) to (149.center);
		\draw [in=90, out=-90, looseness=0.75] (162.center) to (161.center);
		\draw (161.center) to (160.center);
		\draw [in=90, out=-90, looseness=1.25] (154.center) to (163.center);
		\draw [bend right=90] (155.center) to (160.center);
		\draw (162.center) to (168.center);
		\draw (169.center) to (163.center);
		\draw (171.center) to (176.center);
		\draw (176.center) to (177.center);
		\draw (177.center) to (172.center);
		\draw (172.center) to (173.center);
		\draw (173.center) to (174.center);
		\draw (174.center) to (170.center);
		\draw (170.center) to (171.center);
		\draw [in=90, out=-90, looseness=0.75] (182.center) to (181.center);
		\draw (181.center) to (180.center);
		\draw [in=90, out=-90, looseness=1.25] (176.center) to (183.center);
		\draw [bend left=90] (177.center) to (180.center);
		\draw (182.center) to (188.center);
		\draw (189.center) to (183.center);
		\draw (193.center) to (198.center);
		\draw (198.center) to (199.center);
		\draw (199.center) to (194.center);
		\draw (194.center) to (195.center);
		\draw (195.center) to (196.center);
		\draw (196.center) to (192.center);
		\draw (192.center) to (193.center);
		\draw [in=90, out=-90, looseness=0.75] (204.center) to (203.center);
		\draw (203.center) to (202.center);
		\draw [bend left=90] (199.center) to (202.center);
		\draw (204.center) to (210.center);
		\draw [in=90, out=-90, looseness=0.75] (216.center) to (215.center);
		\draw (215.center) to (214.center);
		\draw [bend right=90] (213.center) to (214.center);
		\draw (216.center) to (217.center);
		\draw (198.center) to (218.center);
		\draw (220.center) to (225.center);
		\draw (225.center) to (226.center);
		\draw (226.center) to (221.center);
		\draw (221.center) to (222.center);
		\draw (222.center) to (223.center);
		\draw (223.center) to (219.center);
		\draw (219.center) to (220.center);
		\draw [in=90, out=-90, looseness=0.75] (231.center) to (230.center);
		\draw (230.center) to (229.center);
		\draw [bend left=90] (226.center) to (229.center);
		\draw (231.center) to (236.center);
		\draw [in=90, out=-90, looseness=0.75] (241.center) to (240.center);
		\draw (240.center) to (239.center);
		\draw [bend right=90] (238.center) to (239.center);
		\draw (241.center) to (242.center);
		\draw (225.center) to (243.center);
		\draw (244.center) to (245.center);
		\draw [bend left=90, looseness=0.75] (168.center) to (249.center);
		\draw (249.center) to (250.center);
		\draw (252.center) to (251.center);
		\draw [bend left=90, looseness=0.75] (251.center) to (188.center);
		\draw (254.center) to (253.center);
		\draw (256.center) to (255.center);
		\draw (258.center) to (257.center);
		\draw (260.center) to (259.center);
		\draw [bend left=90, looseness=0.75] (253.center) to (210.center);
		\draw [bend left=90, looseness=0.75] (217.center) to (255.center);
		\draw [bend left=90, looseness=0.75] (257.center) to (236.center);
		\draw [bend left=90, looseness=0.75] (242.center) to (259.center);
		\draw [style=gray] (166.center) to (165.center);
		\draw [style=gray] (165.center) to (164.center);
		\draw [style=gray] (164.center) to (167.center);
		\draw [style=gray] (167.center) to (166.center);
		\draw [style=gray] (187.center) to (184.center);
		\draw [style=gray] (184.center) to (185.center);
		\draw [style=gray] (185.center) to (186.center);
		\draw [style=gray] (186.center) to (187.center);
		\draw [style=gray] (209.center) to (206.center);
		\draw [style=gray] (206.center) to (207.center);
		\draw [style=gray] (207.center) to (208.center);
		\draw [style=gray] (208.center) to (209.center);
		\draw [style=gray] (232.center) to (233.center);
		\draw [style=gray] (233.center) to (234.center);
		\draw [style=gray] (234.center) to (235.center);
		\draw [style=gray] (235.center) to (232.center);
	\end{pgfonlayer}
\end{tikzpicture}
\end{equation}

The \texttt{snake_removal} method prescribes the autonomisation of each word and subsequent yanking of the wires, as done in Def.~2.12 of Ref.~\cite{dunnvicary}:

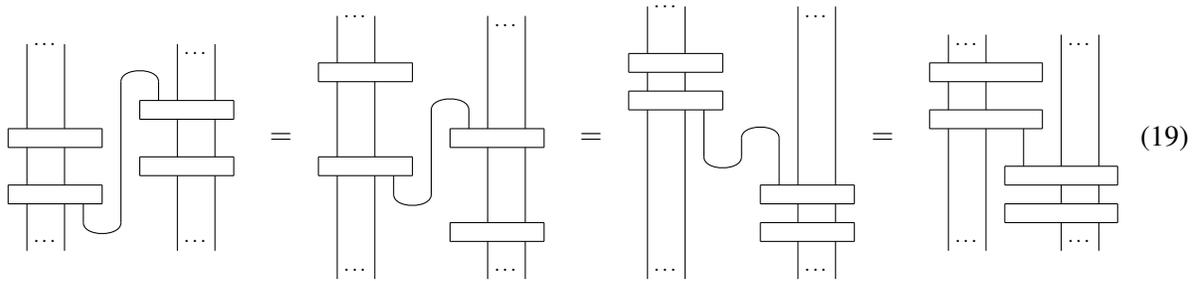
\begin{equation}\label{eq:snakeremovaljamie}
\input{./figures/tikz/Stefano/snake-removal-jamie.tikz}
\end{equation}

\noindent
The end result is a ``snake-free'' diagram with no cups and caps, which can be interpreted any symmetric monoidal category.

For example, consider the autonomisation of the following grammatical sentence:
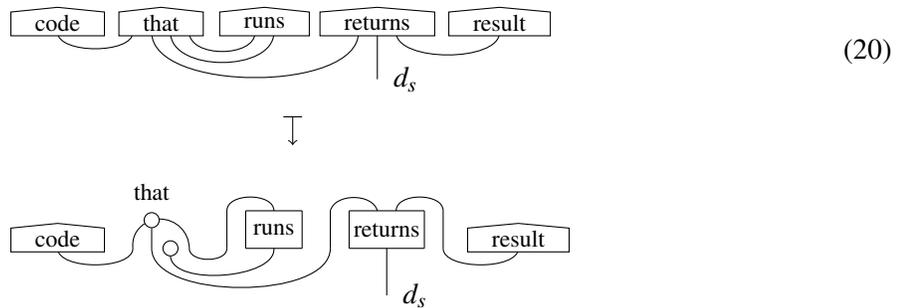
\begin{equation}
    \begin{tikzpicture}
	\begin{pgfonlayer}{nodelayer}
		\node [style=langstate] (2) at (6.25, 0.75) {\footnotesize that};
		\node [style=langstate] (3) at (9, 0.75) {\footnotesize runs};
		\node [style=langstate] (9) at (15.25, 0.75) {\footnotesize result};
		\node [style=langstate] (10) at (12, 0.75) {\footnotesize returns};
		\node [style=none] (13) at (3.5, 0.5) {};
		\node [style=none] (15) at (6, 0.5) {};
		\node [style=none] (16) at (8.75, 0.5) {};
		\node [style=none] (20) at (11.5, 0.5) {};
		\node [style=none] (21) at (15.25, 0.5) {};
		\node [style=none] (23) at (5.5, 0.5) {};
		\node [style=none] (24) at (6.5, 0.5) {};
		\node [style=none] (25) at (7, 0.5) {};
		\node [style=none] (26) at (9.25, 0.5) {};
		\node [style=none] (30) at (12.5, 0.5) {};
		\node [style=none] (31) at (12, 0.5) {};
		\node [style=langstate] (33) at (3.5, 0.75) {\footnotesize code};
		\node [style=none] (34) at (12, -0.75) {};
		\node [style=none] (35) at (12.75, -0.75) {$d_s$};
		\node [style=none] (40) at (3.5, -5.25) {};
		\node [style=none] (41) at (6, -5.25) {};
		\node [style=none] (42) at (8, -5.25) {};
		\node [style=none] (43) at (10.75, -5.25) {};
		\node [style=none] (44) at (15.75, -5.25) {};
		\node [style=none] (45) at (5.5, -5.25) {};
		\node [style=none] (47) at (7, -5.25) {};
		\node [style=none] (48) at (9.25, -5.25) {};
		\node [style=none] (49) at (13.75, -5.25) {};
		\node [style=none] (50) at (12.25, -5.25) {};
		\node [style=langstate] (51) at (3.5, -5) {\footnotesize code};
		\node [style=none] (52) at (12.25, -6.5) {};
		\node [style=none] (53) at (13, -6.5) {$d_s$};
		\node [style=none] (54) at (9.5, -1.75) {};
		\node [style=none] (55) at (9.75, -1.75) {};
		\node [style=none] (56) at (9.75, -1.75) {};
		\node [style=none] (57) at (10, -1.75) {};
		\node [style=none] (58) at (9.75, -2.5) {};
		\node [style=none] (59) at (8, -4.25) {};
		\node [style=white dot] (60) at (6.5, -5.25) {};
		\node [style=white dot] (61) at (6, -4.5) {};
		\node [style=none] (62) at (9.25, -4.25) {};
		\node [style=none] (63) at (6, -3.75) {\footnotesize that};
		\node [style=none] (64) at (8.5, -4.25) {};
		\node [style=none] (65) at (10, -4.25) {};
		\node [style=none] (66) at (8.5, -5.25) {};
		\node [style=none] (67) at (10, -5.25) {};
		\node [style=none] (68) at (9.25, -4.75) {\footnotesize runs};
		\node [style=none] (69) at (12.25, -4.75) {\footnotesize returns};
		\node [style=langstate] (70) at (15.75, -5) {\footnotesize result};
		\node [style=none] (71) at (11.25, -5.25) {};
		\node [style=none] (72) at (13.25, -5.25) {};
		\node [style=none] (73) at (11.25, -4.25) {};
		\node [style=none] (74) at (13.25, -4.25) {};
		\node [style=none] (75) at (10.75, -4.25) {};
		\node [style=none] (76) at (13.75, -4.25) {};
		\node [style=none] (77) at (12, -4.25) {};
		\node [style=none] (78) at (12.5, -4.25) {};
	\end{pgfonlayer}
	\begin{pgfonlayer}{edgelayer}
		\draw [bend right=90] (25.center) to (16.center);
		\draw [bend right=90, looseness=0.75] (30.center) to (21.center);
		\draw [bend right=90, looseness=0.75] (13.center) to (23.center);
		\draw [bend right=90] (24.center) to (26.center);
		\draw [bend right=90, looseness=0.75] (15.center) to (20.center);
		\draw (34.center) to (31.center);
		\draw [bend right=90] (47.center) to (42.center);
		\draw [bend right=90, looseness=0.75] (49.center) to (44.center);
		\draw [bend right=90, looseness=0.75] (40.center) to (45.center);
		\draw [bend right=90, looseness=0.75] (41.center) to (43.center);
		\draw (52.center) to (50.center);
		\draw (54.center) to (57.center);
		\draw [style=diredge] (56.center) to (58.center);
		\draw [bend left=90] (59.center) to (62.center);
		\draw (59.center) to (42.center);
		\draw [bend left=45] (61) to (47.center);
		\draw [bend left] (45.center) to (61);
		\draw (41.center) to (61);
		\draw (64.center) to (65.center);
		\draw (65.center) to (67.center);
		\draw (66.center) to (67.center);
		\draw (64.center) to (66.center);
		\draw [bend left=90, looseness=0.75] (48.center) to (60);
		\draw (73.center) to (74.center);
		\draw (73.center) to (71.center);
		\draw (71.center) to (72.center);
		\draw (72.center) to (74.center);
		\draw (75.center) to (43.center);
		\draw (76.center) to (49.center);
		\draw [bend left=90] (78.center) to (76.center);
		\draw [bend left=75] (75.center) to (77.center);
	\end{pgfonlayer}
\end{tikzpicture}
\end{equation}
(where we have used the classical structure ansatz for "that" from \eqref{eq:capfrobansatz}).
The \texttt{snake_removal} method would result in the following ``snake-free'' diagram :
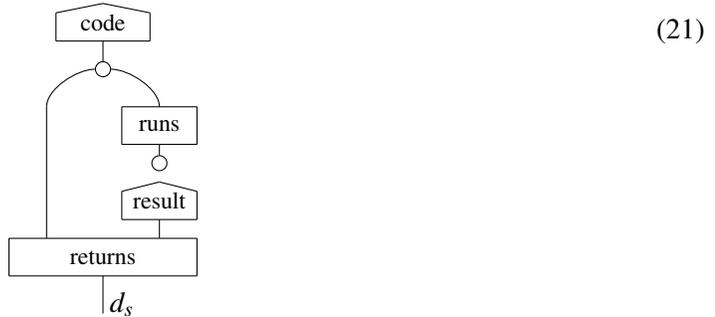
\begin{equation}
    \begin{tikzpicture}
	\begin{pgfonlayer}{nodelayer}
		\node [style=none] (1) at (-3.5, 0.75) {};
		\node [style=none] (2) at (-4.75, 0.5) {};
		\node [style=none] (3) at (-4.75, -0.25) {};
		\node [style=none] (4) at (-2.25, -0.25) {};
		\node [style=none] (5) at (-2.25, 0.5) {};
		\node [style=none] (6) at (-3.5, -0.25) {};
		\node [style=white dot] (7) at (-3.5, -1) {};
		\node [style=none] (8) at (-2, -2) {};
		\node [style=none] (10) at (-3, -2) {};
		\node [style=none] (11) at (-1, -2) {};
		\node [style=none] (12) at (-3, -3) {};
		\node [style=none] (13) at (-1, -3) {};
		\node [style=none] (14) at (-2, -3) {};
		\node [style=white dot] (15) at (-2, -3.5) {};
		\node [style=none] (16) at (-5, -2) {};
		\node [style=none] (17) at (-5, -5.5) {};
		\node [style=none] (18) at (-2, -5.5) {};
		\node [style=none] (19) at (-2, -5) {};
		\node [style=none] (20) at (-3, -5) {};
		\node [style=none] (21) at (-1, -5) {};
		\node [style=none] (22) at (-3, -4.25) {};
		\node [style=none] (23) at (-1, -4.25) {};
		\node [style=none] (25) at (-2, -4) {};
		\node [style=none] (26) at (-6, -5.5) {};
		\node [style=none] (27) at (-6, -6.5) {};
		\node [style=none] (28) at (-1, -6.5) {};
		\node [style=none] (29) at (-1, -5.5) {};
		\node [style=none] (30) at (-3.5, -6.5) {};
		\node [style=none] (31) at (-3.5, -7.5) {};
		\node [style=none] (32) at (-3.5, 0.25) {\footnotesize code};
		\node [style=none] (33) at (-2, -2.5) {\footnotesize runs};
		\node [style=none] (34) at (-2, -4.5) {\footnotesize result};
		\node [style=none] (35) at (-3.5, -6) {\footnotesize returns};
		\node [style=none] (36) at (-3, -7.25) {$d_s$};
	\end{pgfonlayer}
	\begin{pgfonlayer}{edgelayer}
		\draw (26.center) to (27.center);
		\draw (27.center) to (30.center);
		\draw (17.center) to (26.center);
		\draw (17.center) to (18.center);
		\draw (18.center) to (29.center);
		\draw (29.center) to (28.center);
		\draw (28.center) to (30.center);
		\draw (18.center) to (19.center);
		\draw (19.center) to (20.center);
		\draw (20.center) to (22.center);
		\draw (22.center) to (25.center);
		\draw (25.center) to (23.center);
		\draw (23.center) to (21.center);
		\draw (21.center) to (19.center);
		\draw (17.center) to (16.center);
		\draw (10.center) to (12.center);
		\draw (12.center) to (14.center);
		\draw (14.center) to (13.center);
		\draw (13.center) to (11.center);
		\draw [in=90, out=0, looseness=0.75] (7) to (8.center);
		\draw [in=-180, out=90, looseness=0.75] (16.center) to (7);
		\draw (3.center) to (6.center);
		\draw (6.center) to (7);
		\draw (6.center) to (4.center);
		\draw (4.center) to (5.center);
		\draw (1.center) to (5.center);
		\draw (2.center) to (1.center);
		\draw (2.center) to (3.center);
		\draw (10.center) to (11.center);
		\draw (30.center) to (31.center);
		\draw (14.center) to (15);
	\end{pgfonlayer}
\end{tikzpicture}
\end{equation}
Snake removal can also be seen in action in Ref \cite{felice2020functorial}.

When translating the resulting snake-free diagrams into quantum circuits, it is important to note that process-state duality requires \emph{all linear maps} to be available in the autonomisation process, not only the unitary ones.
The realisation of non-unitary maps requires ancillary states and post-selection: this leads to an increase in circuit width and---ceteris paribus---an exponentially higher number of samples required during computation.
If post-selection is not a viable option, then the restriction to unitary maps in turn imposes significant restrictions on the states available for words.
For example, adjectives cannot change the semantic distance between the nouns they modify: a ``spherical cow'' and a ``spherical chicken'' will have the same semantic distance that the unmodified ``cow'' and ``chicken'' previously had, regardless of whether they are in a vacuum or not.\\

\noindent
(A Python implementation of the \texttt{snake_removal} method, as part of the DisCoPy toolbox for monoidal categories, is available at \href{https://github.com/oxford-quantum-group/discopy}{github.com/oxford-quantum-group/discopy}. For more details see the accompanying technical paper, Ref.~\cite{felice2020discopy})

\section{From Diagram to Circuit}
\label{section:diagram2circuit}

The last step in our pipeline is the association of ans\"atze to words, either in the form of state ans\"atze (for the \texttt{bigraph} method) or in the form of process ans\"atze (for the \texttt{snake_removal} method).
We consider two generic families of unitary qubit ans\"atze, the CNOT+U(3) ones and the IQP ones.
Each atomic type $\tau$ is mapped to one or more qubits, i.e. we have $d_\tau = 2^{n_\tau}$.
More general ans\"atze are derived from the unitary ones as follows:

\begin{itemize}
    \itemsep0mm
    \item state ans\"atze are obtained by application of the unitary ans\"atze to the Pauli Z $\ket{0}$ state;
    \item effect ans\"atze are obtained by transposition of the unitary ans\"atze and post-selection onto the Pauli Z measurement outcome corresponding to the $\bra{0}$ effect;
    \footnote{Not that the word post-selection is used here to denote the linear process where no re-normalisation of probabilities is performed.}
    \item more general linear map ans\"atze method are obtained by using ancillary qubits prepared in the $\ket{0}$ state and/or post-selecting onto the measurement outcome corresponding to the $\bra{0}$ effect.
\end{itemize}

\noindent
For the \texttt{bigraph} method, semantics $\hat{\textbf{G}} \rightarrow \textbf{fHilb}$ are given by associating each word to a linear map ansatz and then constructing the following monoidal functor:

\begin{itemize}
    \item word states are mapped to the state ans\"atze described above;
    \item word effects are mapped to the effect ans\"atze described above;
    \item wire crossing are mapped to swaps, cups are mapped to preparation of a Bell state, caps are mapped to post-selection onto the measurement outcome corresponding to the same Bell state.
\end{itemize}

\noindent
For the \texttt{snake_removal} method, semantics $\hat{\textbf{G}}_{aut} \rightarrow \textbf{fHilb}$ are given by associating each word to a linear map ansatz and then constructing the following monoidal functor:
\begin{itemize}
    \item the chosen word representatives in $\hat{\textbf{G}}_{aut}$ are mapped to the linear map ans\"atze;
    \item wire crossing are mapped to swaps;
\end{itemize}

\noindent
With the exception of functional and connection words, the ans\"atze are parametrised.
Typically we associate a single parametric ansatz to all words with the same POS, with the specific values of the parameters distinguishing between the words.

\subsection{CNOT+U(3) ans\"atze}

This family of ans\"atze consists of unitary quantum circuits formed by alternating layers of single-qubits rotations in X and Z with layers of CNOT gates between neighbouring qubits.
The examples below---for 1, 2 and 3 qubits respectively---are written in ZX calculus notation \cite{coecke2008interacting}.
Single-qubits white and black dots are rotations in Pauli Z and Pauli X respectively, while a black and a white dot connected by a horizontal line is a CNOT gate:

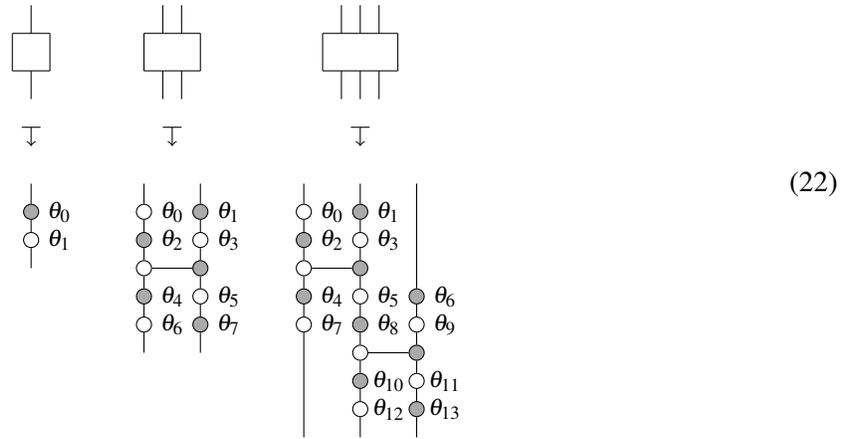
\begin{equation}
\begin{tikzpicture}
	\begin{pgfonlayer}{nodelayer}
		\node [style=none] (0) at (-8.5, 4) {};
		\node [style=none] (1) at (-7.5, 4) {};
		\node [style=none] (2) at (-8.5, 3) {};
		\node [style=none] (3) at (-7.5, 3) {};
		\node [style=none] (4) at (-8, 4.75) {};
		\node [style=none] (5) at (-8, 2.25) {};
		\node [style=none] (6) at (-8, 4) {};
		\node [style=none] (7) at (-8, 3) {};
		\node [style=none] (8) at (-5, 4) {};
		\node [style=none] (9) at (-3.5, 4) {};
		\node [style=none] (10) at (-5, 3) {};
		\node [style=none] (11) at (-3.5, 3) {};
		\node [style=none] (12) at (-4.5, 4.75) {};
		\node [style=none] (15) at (-4.5, 3) {};
		\node [style=none] (16) at (-4, 3) {};
		\node [style=none] (17) at (-4, 4.75) {};
		\node [style=none] (18) at (-4.5, 4) {};
		\node [style=none] (19) at (-4, 4) {};
		\node [style=none] (20) at (-4.5, 2.25) {};
		\node [style=none] (21) at (-4, 2.25) {};
		\node [style=none] (22) at (-0.25, 4) {};
		\node [style=none] (23) at (1.25, 4) {};
		\node [style=none] (24) at (-0.25, 3) {};
		\node [style=none] (25) at (1.25, 3) {};
		\node [style=none] (26) at (0.25, 4.75) {};
		\node [style=none] (29) at (0.75, 4.75) {};
		\node [style=none] (32) at (0.25, 2.25) {};
		\node [style=none] (33) at (0.75, 2.25) {};
		\node [style=none] (34) at (0.25, 4) {};
		\node [style=none] (35) at (0.75, 4) {};
		\node [style=none] (36) at (0.25, 3) {};
		\node [style=none] (37) at (0.75, 3) {};
		\node [style=none] (38) at (1.75, 4) {};
		\node [style=none] (39) at (1.75, 3) {};
		\node [style=none] (40) at (1.25, 2.25) {};
		\node [style=none] (41) at (1.25, 4.75) {};
		\node [style=none] (42) at (-8.25, 1.5) {};
		\node [style=none] (43) at (-7.75, 1.5) {};
		\node [style=none] (44) at (-8, 1.5) {};
		\node [style=none] (45) at (-8, 1) {};
		\node [style=none] (46) at (-4.5, 1.5) {};
		\node [style=none] (47) at (-4, 1.5) {};
		\node [style=none] (48) at (-4.25, 1.5) {};
		\node [style=none] (49) at (-4.25, 1) {};
		\node [style=none] (50) at (0.5, 1.5) {};
		\node [style=none] (51) at (1, 1.5) {};
		\node [style=none] (52) at (0.75, 1.5) {};
		\node [style=none] (53) at (0.75, 1) {};
		\node [style=white dot] (54) at (-8, -1.5) {};
		\node [style=gray dot] (55) at (-8, -0.75) {};
		\node [style=none] (56) at (-8, 0) {};
		\node [style=none] (57) at (-8, -2.25) {};
		\node [style=white dot] (58) at (-5, -0.75) {};
		\node [style=gray dot] (59) at (-5, -1.5) {};
		\node [style=gray dot] (60) at (-3.5, -0.75) {};
		\node [style=white dot] (61) at (-3.5, -1.5) {};
		\node [style=white dot] (62) at (-5, -2.25) {};
		\node [style=gray dot] (63) at (-5, -3) {};
		\node [style=gray dot] (64) at (-3.5, -2.25) {};
		\node [style=white dot] (65) at (-3.5, -3) {};
		\node [style=white dot] (66) at (-5, -3.75) {};
		\node [style=gray dot] (67) at (-3.5, -3.75) {};
		\node [style=none] (68) at (-5, 0) {};
		\node [style=none] (69) at (-3.5, 0) {};
		\node [style=none] (70) at (-5, -4.5) {};
		\node [style=none] (71) at (-3.5, -4.5) {};
		\node [style=white dot] (72) at (-0.75, -0.75) {};
		\node [style=gray dot] (73) at (-0.75, -1.5) {};
		\node [style=gray dot] (74) at (0.75, -0.75) {};
		\node [style=white dot] (75) at (0.75, -1.5) {};
		\node [style=white dot] (76) at (-0.75, -2.25) {};
		\node [style=gray dot] (77) at (-0.75, -3) {};
		\node [style=gray dot] (78) at (0.75, -2.25) {};
		\node [style=white dot] (79) at (0.75, -3) {};
		\node [style=white dot] (80) at (-0.75, -3.75) {};
		\node [style=gray dot] (81) at (0.75, -3.75) {};
		\node [style=none] (82) at (-0.75, 0) {};
		\node [style=none] (83) at (0.75, 0) {};
		\node [style=none] (84) at (-0.75, -6.75) {};
		\node [style=white dot] (85) at (0.75, -4.5) {};
		\node [style=gray dot] (86) at (2.25, -4.5) {};
		\node [style=white dot] (87) at (2.25, -3.75) {};
		\node [style=gray dot] (88) at (2.25, -3) {};
		\node [style=gray dot] (89) at (0.75, -5.25) {};
		\node [style=white dot] (90) at (2.25, -5.25) {};
		\node [style=white dot] (91) at (0.75, -6) {};
		\node [style=gray dot] (92) at (2.25, -6) {};
		\node [style=none] (93) at (2.25, 0) {};
		\node [style=none] (94) at (2.25, -6.75) {};
		\node [style=none] (95) at (0.75, -6.75) {};
		\node [style=none] (96) at (-7.25, -0.75) {\footnotesize $\theta_0$};
		\node [style=none] (97) at (-7.25, -1.5) {\footnotesize $\theta_1$};
		\node [style=none] (98) at (-4.25, -0.75) {\footnotesize $\theta_0$};
		\node [style=none] (99) at (-2.75, -0.75) {\footnotesize $\theta_1$};
		\node [style=none] (100) at (-4.25, -1.5) {\footnotesize $\theta_2$};
		\node [style=none] (101) at (-2.75, -1.5) {\footnotesize $\theta_3$};
		\node [style=none] (102) at (-4.25, -3) {\footnotesize $\theta_4$};
		\node [style=none] (103) at (-2.75, -3) {\footnotesize $\theta_5$};
		\node [style=none] (104) at (-4.25, -3.75) {\footnotesize $\theta_6$};
		\node [style=none] (105) at (-2.75, -3.75) {\footnotesize $\theta_7$};
		\node [style=none] (106) at (0, -0.75) {\footnotesize $\theta_0$};
		\node [style=none] (107) at (1.5, -0.75) {\footnotesize $\theta_1$};
		\node [style=none] (108) at (0, -1.5) {\footnotesize $\theta_2$};
		\node [style=none] (109) at (1.5, -1.5) {\footnotesize $\theta_3$};
		\node [style=none] (110) at (0, -3) {\footnotesize $\theta_4$};
		\node [style=none] (111) at (1.5, -3) {\footnotesize $\theta_5$};
		\node [style=none] (112) at (0, -3.75) {\footnotesize $\theta_7$};
		\node [style=none] (113) at (1.5, -3.75) {\footnotesize $\theta_8$};
		\node [style=none] (114) at (1.5, -5.25) {\footnotesize $\theta_{10}$};
		\node [style=none] (115) at (3, -5.25) {\footnotesize $\theta_{11}$};
		\node [style=none] (116) at (1.5, -6) {\footnotesize $\theta_{12}$};
		\node [style=none] (117) at (3, -6) {\footnotesize $\theta_{13}$};
		\node [style=none] (118) at (3, -3) {\footnotesize $\theta_6$};
		\node [style=none] (119) at (3, -3.75) {\footnotesize $\theta_9$};
	\end{pgfonlayer}
	\begin{pgfonlayer}{edgelayer}
		\draw (0.center) to (6.center);
		\draw (6.center) to (1.center);
		\draw (1.center) to (3.center);
		\draw (3.center) to (7.center);
		\draw (7.center) to (2.center);
		\draw (2.center) to (0.center);
		\draw (6.center) to (4.center);
		\draw (7.center) to (5.center);
		\draw (9.center) to (11.center);
		\draw (11.center) to (15.center);
		\draw (15.center) to (10.center);
		\draw (10.center) to (8.center);
		\draw (8.center) to (18.center);
		\draw (18.center) to (19.center);
		\draw (19.center) to (9.center);
		\draw (12.center) to (18.center);
		\draw (19.center) to (17.center);
		\draw (15.center) to (20.center);
		\draw (16.center) to (21.center);
		\draw (24.center) to (22.center);
		\draw (22.center) to (34.center);
		\draw (34.center) to (35.center);
		\draw (35.center) to (23.center);
		\draw (23.center) to (38.center);
		\draw (38.center) to (39.center);
		\draw (39.center) to (25.center);
		\draw (25.center) to (37.center);
		\draw (37.center) to (36.center);
		\draw (36.center) to (24.center);
		\draw (26.center) to (34.center);
		\draw (35.center) to (29.center);
		\draw (41.center) to (23.center);
		\draw (36.center) to (32.center);
		\draw (33.center) to (37.center);
		\draw (25.center) to (40.center);
		\draw (42.center) to (44.center);
		\draw (44.center) to (43.center);
		\draw [style=diredge] (44.center) to (45.center);
		\draw (46.center) to (48.center);
		\draw (48.center) to (47.center);
		\draw [style=diredge] (48.center) to (49.center);
		\draw (50.center) to (52.center);
		\draw (52.center) to (51.center);
		\draw [style=diredge] (52.center) to (53.center);
		\draw (56.center) to (55);
		\draw (55) to (54);
		\draw (54) to (57.center);
		\draw (68.center) to (58);
		\draw (69.center) to (60);
		\draw (58) to (59);
		\draw (60) to (61);
		\draw (61) to (64);
		\draw (64) to (65);
		\draw (65) to (67);
		\draw (67) to (71.center);
		\draw (70.center) to (66);
		\draw (66) to (63);
		\draw (63) to (62);
		\draw (62) to (59);
		\draw (62) to (64);
		\draw (82.center) to (72);
		\draw (83.center) to (74);
		\draw (72) to (73);
		\draw (74) to (75);
		\draw (75) to (78);
		\draw (78) to (79);
		\draw (79) to (81);
		\draw (84.center) to (80);
		\draw (80) to (77);
		\draw (77) to (76);
		\draw (76) to (73);
		\draw (76) to (78);
		\draw (85) to (86);
		\draw (88) to (93.center);
		\draw (88) to (87);
		\draw (87) to (86);
		\draw (86) to (90);
		\draw (90) to (92);
		\draw (92) to (94.center);
		\draw (91) to (95.center);
		\draw (91) to (89);
		\draw (89) to (85);
		\draw (85) to (81);
	\end{pgfonlayer}
\end{tikzpicture}
\end{equation}

\noindent
State ans\"atze are obtained by applying the unitary ans\"atze to the zero state of the computational basis, following the convention on IBMQ devices,
and
effect ans\"atze are obtained by transposing the state ans\"atze in the computational basis:

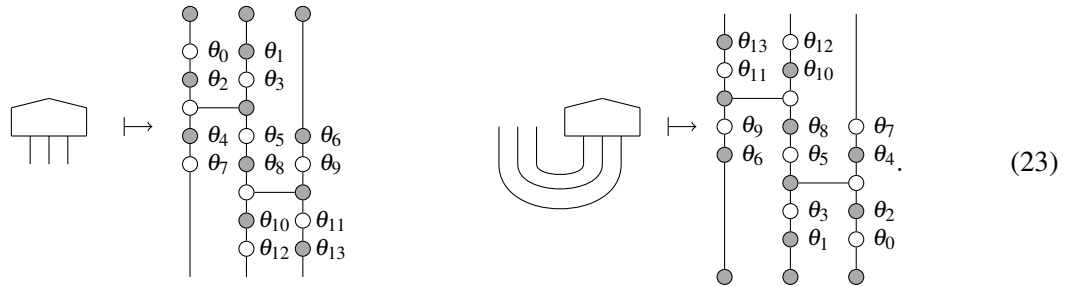
\begin{equation}
\begin{tikzpicture}
	\begin{pgfonlayer}{nodelayer}
		\node [style=none] (22) at (-3, 1.5) {};
		\node [style=none] (24) at (-3, 0.75) {};
		\node [style=none] (25) at (-1.5, 0.75) {};
		\node [style=none] (32) at (-2.5, 0) {};
		\node [style=none] (33) at (-2, 0) {};
		\node [style=none] (35) at (-2, 1.75) {};
		\node [style=none] (36) at (-2.5, 0.75) {};
		\node [style=none] (37) at (-2, 0.75) {};
		\node [style=none] (38) at (-1, 1.5) {};
		\node [style=none] (39) at (-1, 0.75) {};
		\node [style=none] (40) at (-1.5, 0) {};
		\node [style=none] (53) at (0.75, 1) {};
		\node [style=white dot] (72) at (1.75, 3) {};
		\node [style=gray dot] (73) at (1.75, 2.25) {};
		\node [style=gray dot] (74) at (3.25, 3) {};
		\node [style=white dot] (75) at (3.25, 2.25) {};
		\node [style=white dot] (76) at (1.75, 1.5) {};
		\node [style=gray dot] (77) at (1.75, 0.75) {};
		\node [style=gray dot] (78) at (3.25, 1.5) {};
		\node [style=white dot] (79) at (3.25, 0.75) {};
		\node [style=white dot] (80) at (1.75, 0) {};
		\node [style=gray dot] (81) at (3.25, 0) {};
		\node [style=none] (84) at (1.75, -3) {};
		\node [style=white dot] (85) at (3.25, -0.75) {};
		\node [style=gray dot] (86) at (4.75, -0.75) {};
		\node [style=white dot] (87) at (4.75, 0) {};
		\node [style=gray dot] (88) at (4.75, 0.75) {};
		\node [style=gray dot] (89) at (3.25, -1.5) {};
		\node [style=white dot] (90) at (4.75, -1.5) {};
		\node [style=white dot] (91) at (3.25, -2.25) {};
		\node [style=gray dot] (92) at (4.75, -2.25) {};
		\node [style=none] (94) at (4.75, -3) {};
		\node [style=none] (95) at (3.25, -3) {};
		\node [style=none] (106) at (2.5, 3) {\footnotesize $\theta_0$};
		\node [style=none] (107) at (4, 3) {\footnotesize $\theta_1$};
		\node [style=none] (108) at (2.5, 2.25) {\footnotesize $\theta_2$};
		\node [style=none] (109) at (4, 2.25) {\footnotesize $\theta_3$};
		\node [style=none] (110) at (2.5, 0.75) {\footnotesize $\theta_4$};
		\node [style=none] (111) at (4, 0.75) {\footnotesize $\theta_5$};
		\node [style=none] (112) at (2.5, 0) {\footnotesize $\theta_7$};
		\node [style=none] (113) at (4, 0) {\footnotesize $\theta_8$};
		\node [style=none] (114) at (4, -1.5) {\footnotesize $\theta_{10}$};
		\node [style=none] (115) at (5.5, -1.5) {\footnotesize $\theta_{11}$};
		\node [style=none] (116) at (4, -2.25) {\footnotesize $\theta_{12}$};
		\node [style=none] (117) at (5.5, -2.25) {\footnotesize $\theta_{13}$};
		\node [style=none] (118) at (5.5, 0.75) {\footnotesize $\theta_6$};
		\node [style=none] (119) at (5.5, 0) {\footnotesize $\theta_9$};
		\node [style=none] (120) at (0, 1.25) {};
		\node [style=none] (121) at (0, 0.75) {};
		\node [style=none] (122) at (0, 1) {};
		\node [style=gray dot] (123) at (1.75, 4) {};
		\node [style=gray dot] (124) at (3.25, 4) {};
		\node [style=gray dot] (125) at (4.75, 4) {};
	\end{pgfonlayer}
	\begin{pgfonlayer}{edgelayer}
		\draw (24.center) to (22.center);
		\draw (38.center) to (39.center);
		\draw (39.center) to (25.center);
		\draw (25.center) to (37.center);
		\draw (37.center) to (36.center);
		\draw (36.center) to (24.center);
		\draw (36.center) to (32.center);
		\draw (33.center) to (37.center);
		\draw (25.center) to (40.center);
		\draw (72) to (73);
		\draw (74) to (75);
		\draw (75) to (78);
		\draw (78) to (79);
		\draw (79) to (81);
		\draw (84.center) to (80);
		\draw (80) to (77);
		\draw (77) to (76);
		\draw (76) to (73);
		\draw (76) to (78);
		\draw (85) to (86);
		\draw (88) to (87);
		\draw (87) to (86);
		\draw (86) to (90);
		\draw (90) to (92);
		\draw (92) to (94.center);
		\draw (91) to (95.center);
		\draw (91) to (89);
		\draw (89) to (85);
		\draw (85) to (81);
		\draw (22.center) to (35.center);
		\draw (35.center) to (38.center);
		\draw (120.center) to (122.center);
		\draw (121.center) to (122.center);
		\draw [style=diredge] (122.center) to (53.center);
		\draw (123) to (72);
		\draw (124) to (74);
		\draw (125) to (88);
	\end{pgfonlayer}
\end{tikzpicture}
\hspace{2cm}
\begin{tikzpicture}
	\begin{pgfonlayer}{nodelayer}
		\node [style=none] (22) at (-2.75, 1.5) {};
		\node [style=none] (24) at (-2.75, 0.75) {};
		\node [style=none] (25) at (-1.25, 0.75) {};
		\node [style=none] (32) at (-2.25, 0) {};
		\node [style=none] (33) at (-1.75, 0) {};
		\node [style=none] (35) at (-1.75, 1.75) {};
		\node [style=none] (36) at (-2.25, 0.75) {};
		\node [style=none] (37) at (-1.75, 0.75) {};
		\node [style=none] (38) at (-0.75, 1.5) {};
		\node [style=none] (39) at (-0.75, 0.75) {};
		\node [style=none] (40) at (-1.25, 0) {};
		\node [style=none] (53) at (0.75, 1) {};
		\node [style=white dot] (72) at (5, -2) {};
		\node [style=gray dot] (73) at (5, -1.25) {};
		\node [style=gray dot] (74) at (3.25, -2) {};
		\node [style=white dot] (75) at (3.25, -1.25) {};
		\node [style=white dot] (76) at (5, -0.5) {};
		\node [style=gray dot] (77) at (5, 0.25) {};
		\node [style=gray dot] (78) at (3.25, -0.5) {};
		\node [style=white dot] (79) at (3.25, 0.25) {};
		\node [style=white dot] (80) at (5, 1) {};
		\node [style=gray dot] (81) at (3.25, 1) {};
		\node [style=none] (84) at (5, 4) {};
		\node [style=white dot] (85) at (3.25, 1.75) {};
		\node [style=gray dot] (86) at (1.5, 1.75) {};
		\node [style=white dot] (87) at (1.5, 1) {};
		\node [style=gray dot] (88) at (1.5, 0.25) {};
		\node [style=gray dot] (89) at (3.25, 2.5) {};
		\node [style=white dot] (90) at (1.5, 2.5) {};
		\node [style=white dot] (91) at (3.25, 3.25) {};
		\node [style=gray dot] (92) at (1.5, 3.25) {};
		\node [style=none] (94) at (1.5, 4) {};
		\node [style=none] (95) at (3.25, 4) {};
		\node [style=none] (106) at (5.75, -2) {\footnotesize $\theta_0$};
		\node [style=none] (107) at (4, -2) {\footnotesize $\theta_1$};
		\node [style=none] (108) at (5.75, -1.25) {\footnotesize $\theta_2$};
		\node [style=none] (109) at (4, -1.25) {\footnotesize $\theta_3$};
		\node [style=none] (110) at (5.75, 0.25) {\footnotesize $\theta_4$};
		\node [style=none] (111) at (4, 0.25) {\footnotesize $\theta_5$};
		\node [style=none] (112) at (5.75, 1) {\footnotesize $\theta_7$};
		\node [style=none] (113) at (4, 1) {\footnotesize $\theta_8$};
		\node [style=none] (114) at (4, 2.5) {\footnotesize $\theta_{10}$};
		\node [style=none] (115) at (2.25, 2.5) {\footnotesize $\theta_{11}$};
		\node [style=none] (116) at (4, 3.25) {\footnotesize $\theta_{12}$};
		\node [style=none] (117) at (2.25, 3.25) {\footnotesize $\theta_{13}$};
		\node [style=none] (118) at (2.25, 0.25) {\footnotesize $\theta_6$};
		\node [style=none] (119) at (2.25, 1) {\footnotesize $\theta_9$};
		\node [style=none] (120) at (0, 1.25) {};
		\node [style=none] (121) at (0, 0.75) {};
		\node [style=none] (122) at (0, 1) {};
		\node [style=gray dot] (123) at (5, -3) {};
		\node [style=gray dot] (124) at (3.25, -3) {};
		\node [style=gray dot] (125) at (1.5, -3) {};
		\node [style=none] (126) at (-4.5, 0) {};
		\node [style=none] (127) at (-4, 0) {};
		\node [style=none] (128) at (-3.5, 0) {};
		\node [style=none] (129) at (-4.5, 1) {};
		\node [style=none] (130) at (-4, 1) {};
		\node [style=none] (131) at (-3.5, 1) {};
	\end{pgfonlayer}
	\begin{pgfonlayer}{edgelayer}
		\draw (24.center) to (22.center);
		\draw (38.center) to (39.center);
		\draw (39.center) to (25.center);
		\draw (25.center) to (37.center);
		\draw (37.center) to (36.center);
		\draw (36.center) to (24.center);
		\draw (36.center) to (32.center);
		\draw (33.center) to (37.center);
		\draw (25.center) to (40.center);
		\draw (72) to (73);
		\draw (74) to (75);
		\draw (75) to (78);
		\draw (78) to (79);
		\draw (79) to (81);
		\draw (84.center) to (80);
		\draw (80) to (77);
		\draw (77) to (76);
		\draw (76) to (73);
		\draw (76) to (78);
		\draw (85) to (86);
		\draw (88) to (87);
		\draw (87) to (86);
		\draw (86) to (90);
		\draw (90) to (92);
		\draw (92) to (94.center);
		\draw (91) to (95.center);
		\draw (91) to (89);
		\draw (89) to (85);
		\draw (85) to (81);
		\draw (22.center) to (35.center);
		\draw (35.center) to (38.center);
		\draw (120.center) to (122.center);
		\draw (121.center) to (122.center);
		\draw [style=diredge] (122.center) to (53.center);
		\draw (123) to (72);
		\draw (124) to (74);
		\draw (125) to (88);
		\draw (126.center) to (129.center);
		\draw (130.center) to (127.center);
		\draw (128.center) to (131.center);
		\draw [bend right=90, looseness=1.25] (126.center) to (40.center);
		\draw [bend right=90] (127.center) to (33.center);
		\draw [bend left=90, looseness=0.75] (32.center) to (128.center);
	\end{pgfonlayer}
\end{tikzpicture} .
\end{equation}

\noindent
The effect is post-selection (without re-normalisation) onto the Pauli Z measurement outcome corresponding to the $\bra{0}$ effect.
This family of ans\"atze transforms nicely under reversal of all inputs/outputs, as shown by the following 3-qubit example:

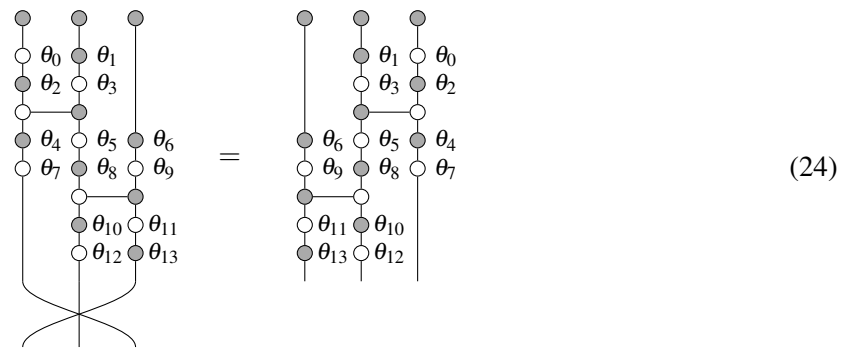
\begin{equation}\label{eq:ansatzReversed}
\begin{tikzpicture}
	\begin{pgfonlayer}{nodelayer}
		\node [style=white dot] (72) at (-3.75, 3) {};
		\node [style=gray dot] (73) at (-3.75, 2.25) {};
		\node [style=gray dot] (74) at (-2.25, 3) {};
		\node [style=white dot] (75) at (-2.25, 2.25) {};
		\node [style=white dot] (76) at (-3.75, 1.5) {};
		\node [style=gray dot] (77) at (-3.75, 0.75) {};
		\node [style=gray dot] (78) at (-2.25, 1.5) {};
		\node [style=white dot] (79) at (-2.25, 0.75) {};
		\node [style=white dot] (80) at (-3.75, 0) {};
		\node [style=gray dot] (81) at (-2.25, 0) {};
		\node [style=none] (84) at (-3.75, -3) {};
		\node [style=white dot] (85) at (-2.25, -0.75) {};
		\node [style=gray dot] (86) at (-0.75, -0.75) {};
		\node [style=white dot] (87) at (-0.75, 0) {};
		\node [style=gray dot] (88) at (-0.75, 0.75) {};
		\node [style=gray dot] (89) at (-2.25, -1.5) {};
		\node [style=white dot] (90) at (-0.75, -1.5) {};
		\node [style=white dot] (91) at (-2.25, -2.25) {};
		\node [style=gray dot] (92) at (-0.75, -2.25) {};
		\node [style=none] (94) at (-0.75, -3) {};
		\node [style=none] (95) at (-2.25, -3) {};
		\node [style=none] (106) at (-3, 3) {\footnotesize $\theta_0$};
		\node [style=none] (107) at (-1.5, 3) {\footnotesize $\theta_1$};
		\node [style=none] (108) at (-3, 2.25) {\footnotesize $\theta_2$};
		\node [style=none] (109) at (-1.5, 2.25) {\footnotesize $\theta_3$};
		\node [style=none] (110) at (-3, 0.75) {\footnotesize $\theta_4$};
		\node [style=none] (111) at (-1.5, 0.75) {\footnotesize $\theta_5$};
		\node [style=none] (112) at (-3, 0) {\footnotesize $\theta_7$};
		\node [style=none] (113) at (-1.5, 0) {\footnotesize $\theta_8$};
		\node [style=none] (114) at (-1.5, -1.5) {\footnotesize $\theta_{10}$};
		\node [style=none] (115) at (0, -1.5) {\footnotesize $\theta_{11}$};
		\node [style=none] (116) at (-1.5, -2.25) {\footnotesize $\theta_{12}$};
		\node [style=none] (117) at (0, -2.25) {\footnotesize $\theta_{13}$};
		\node [style=none] (118) at (0, 0.75) {\footnotesize $\theta_6$};
		\node [style=none] (119) at (0, 0) {\footnotesize $\theta_9$};
		\node [style=gray dot] (123) at (-3.75, 4) {};
		\node [style=gray dot] (124) at (-2.25, 4) {};
		\node [style=gray dot] (125) at (-0.75, 4) {};
		\node [style=white dot] (126) at (6.75, 3) {};
		\node [style=gray dot] (127) at (6.75, 2.25) {};
		\node [style=gray dot] (128) at (5.25, 3) {};
		\node [style=white dot] (129) at (5.25, 2.25) {};
		\node [style=white dot] (130) at (6.75, 1.5) {};
		\node [style=gray dot] (131) at (6.75, 0.75) {};
		\node [style=gray dot] (132) at (5.25, 1.5) {};
		\node [style=white dot] (133) at (5.25, 0.75) {};
		\node [style=white dot] (134) at (6.75, 0) {};
		\node [style=gray dot] (135) at (5.25, 0) {};
		\node [style=none] (136) at (6.75, -3) {};
		\node [style=white dot] (137) at (5.25, -0.75) {};
		\node [style=gray dot] (138) at (3.75, -0.75) {};
		\node [style=white dot] (139) at (3.75, 0) {};
		\node [style=gray dot] (140) at (3.75, 0.75) {};
		\node [style=gray dot] (141) at (5.25, -1.5) {};
		\node [style=white dot] (142) at (3.75, -1.5) {};
		\node [style=white dot] (143) at (5.25, -2.25) {};
		\node [style=gray dot] (144) at (3.75, -2.25) {};
		\node [style=none] (145) at (3.75, -3) {};
		\node [style=none] (146) at (5.25, -3) {};
		\node [style=none] (147) at (7.5, 3) {\footnotesize $\theta_0$};
		\node [style=none] (148) at (6, 3) {\footnotesize $\theta_1$};
		\node [style=none] (149) at (7.5, 2.25) {\footnotesize $\theta_2$};
		\node [style=none] (150) at (6, 2.25) {\footnotesize $\theta_3$};
		\node [style=none] (151) at (7.5, 0.75) {\footnotesize $\theta_4$};
		\node [style=none] (152) at (6, 0.75) {\footnotesize $\theta_5$};
		\node [style=none] (153) at (7.5, 0) {\footnotesize $\theta_7$};
		\node [style=none] (154) at (6, 0) {\footnotesize $\theta_8$};
		\node [style=none] (155) at (6, -1.5) {\footnotesize $\theta_{10}$};
		\node [style=none] (156) at (4.5, -1.5) {\footnotesize $\theta_{11}$};
		\node [style=none] (157) at (6, -2.25) {\footnotesize $\theta_{12}$};
		\node [style=none] (158) at (4.5, -2.25) {\footnotesize $\theta_{13}$};
		\node [style=none] (159) at (4.5, 0.75) {\footnotesize $\theta_6$};
		\node [style=none] (160) at (4.5, 0) {\footnotesize $\theta_9$};
		\node [style=gray dot] (161) at (6.75, 4) {};
		\node [style=gray dot] (162) at (5.25, 4) {};
		\node [style=gray dot] (163) at (3.75, 4) {};
		\node [style=none] (164) at (1.75, 0.25) {$=$};
		\node [style=none] (165) at (-3.75, -4.75) {};
		\node [style=none] (166) at (-0.75, -4.75) {};
		\node [style=none] (167) at (-2.25, -4.75) {};
	\end{pgfonlayer}
	\begin{pgfonlayer}{edgelayer}
		\draw (72) to (73);
		\draw (74) to (75);
		\draw (75) to (78);
		\draw (78) to (79);
		\draw (79) to (81);
		\draw (84.center) to (80);
		\draw (80) to (77);
		\draw (77) to (76);
		\draw (76) to (73);
		\draw (76) to (78);
		\draw (85) to (86);
		\draw (88) to (87);
		\draw (87) to (86);
		\draw (86) to (90);
		\draw (90) to (92);
		\draw (92) to (94.center);
		\draw (91) to (95.center);
		\draw (91) to (89);
		\draw (89) to (85);
		\draw (85) to (81);
		\draw (123) to (72);
		\draw (124) to (74);
		\draw (125) to (88);
		\draw (126) to (127);
		\draw (128) to (129);
		\draw (129) to (132);
		\draw (132) to (133);
		\draw (133) to (135);
		\draw (136.center) to (134);
		\draw (134) to (131);
		\draw (131) to (130);
		\draw (130) to (127);
		\draw (130) to (132);
		\draw (137) to (138);
		\draw (140) to (139);
		\draw (139) to (138);
		\draw (138) to (142);
		\draw (142) to (144);
		\draw (144) to (145.center);
		\draw (143) to (146.center);
		\draw (143) to (141);
		\draw (141) to (137);
		\draw (137) to (135);
		\draw (161) to (126);
		\draw (162) to (128);
		\draw (163) to (140);
		\draw [in=90, out=-90, looseness=0.50] (84.center) to (166.center);
		\draw [in=90, out=-90, looseness=0.50] (94.center) to (165.center);
		\draw (95.center) to (167.center);
	\end{pgfonlayer}
\end{tikzpicture}
\end{equation}

We have said before that functional and connection words are often modelled in the DisCoCat literature using spiders \cite{frobeniusanatomy,Sadrzadeh2013,Sadrzadeh2014}.
As a consequence, it is interesting to see how spiders can be realised as non-parametric instances of CNOT+U(3) ans\"atze.
Spiders with the same number of input and output legs are obtained from alternating CNOT-TONC ladders with preparation and post-selection on ancillary qubits:

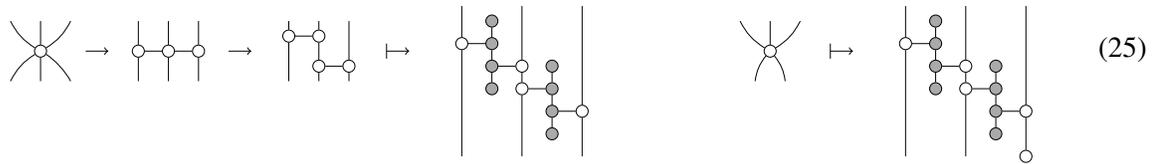
\begin{equation}
\scalebox{0.8}{$\begin{tikzpicture}
	\begin{pgfonlayer}{nodelayer}
		\node [style=white dot] (0) at (-5, 0) {};
		\node [style=none] (1) at (-6, 1) {};
		\node [style=none] (2) at (-5, 1) {};
		\node [style=none] (3) at (-4, 1) {};
		\node [style=none] (4) at (-5, -1) {};
		\node [style=none] (5) at (-6, -1) {};
		\node [style=none] (6) at (-4, -1) {};
		\node [style=none] (8) at (-1.75, 1) {};
		\node [style=none] (9) at (-0.75, 1) {};
		\node [style=none] (10) at (0.25, 1) {};
		\node [style=none] (11) at (-0.75, -1) {};
		\node [style=none] (12) at (-1.75, -1) {};
		\node [style=none] (13) at (0.25, -1) {};
		\node [style=white dot] (14) at (-1.75, 0) {};
		\node [style=white dot] (15) at (-0.75, 0) {};
		\node [style=white dot] (16) at (0.25, 0) {};
		\node [style=none] (17) at (-3.5, 0) {};
		\node [style=none] (18) at (-2.75, 0) {};
		\node [style=none] (19) at (3.25, 1) {};
		\node [style=none] (20) at (4.25, 1) {};
		\node [style=none] (21) at (5.25, 1) {};
		\node [style=none] (23) at (3.25, -1) {};
		\node [style=none] (24) at (5.25, -1) {};
		\node [style=white dot] (25) at (3.25, 0.5) {};
		\node [style=white dot] (26) at (4.25, 0.5) {};
		\node [style=none] (28) at (1.25, 0) {};
		\node [style=none] (29) at (2, 0) {};
		\node [style=white dot] (30) at (4.25, -0.5) {};
		\node [style=white dot] (31) at (5.25, -0.5) {};
		\node [style=none] (32) at (4.25, -1) {};
		\node [style=none] (33) at (9, 1.5) {};
		\node [style=none] (36) at (9, -3.5) {};
		\node [style=white dot] (38) at (9, 0.25) {};
		\node [style=none] (43) at (6.5, 0) {};
		\node [style=none] (44) at (7.25, 0) {};
		\node [style=none] (45) at (6.5, 0.25) {};
		\node [style=none] (46) at (6.5, -0.25) {};
		\node [style=gray dot] (47) at (10, 1) {};
		\node [style=gray dot] (48) at (10, 0.25) {};
		\node [style=gray dot] (49) at (10, -0.5) {};
		\node [style=gray dot] (50) at (10, -1.25) {};
		\node [style=white dot] (51) at (11, -0.5) {};
		\node [style=white dot] (52) at (11, -1.25) {};
		\node [style=gray dot] (53) at (12, -0.5) {};
		\node [style=gray dot] (54) at (12, -1.25) {};
		\node [style=gray dot] (55) at (12, -2) {};
		\node [style=gray dot] (56) at (12, -2.75) {};
		\node [style=white dot] (57) at (13, -2) {};
		\node [style=none] (58) at (13, -3.5) {};
		\node [style=none] (59) at (11, -3.5) {};
		\node [style=none] (60) at (11, 1.5) {};
		\node [style=none] (61) at (13, 1.5) {};
	\end{pgfonlayer}
	\begin{pgfonlayer}{edgelayer}
		\draw [bend right=15] (1.center) to (0);
		\draw (0) to (2.center);
		\draw [bend right=15] (0) to (3.center);
		\draw [bend right=15] (0) to (5.center);
		\draw (0) to (4.center);
		\draw [bend left=15, looseness=0.75] (0) to (6.center);
		\draw (14) to (8.center);
		\draw (14) to (12.center);
		\draw (14) to (15);
		\draw (15) to (9.center);
		\draw (15) to (16);
		\draw (16) to (13.center);
		\draw (16) to (10.center);
		\draw (15) to (11.center);
		\draw [style=diredge] (17.center) to (18.center);
		\draw (25) to (19.center);
		\draw (25) to (23.center);
		\draw (25) to (26);
		\draw (26) to (20.center);
		\draw [style=diredge] (28.center) to (29.center);
		\draw (24.center) to (31);
		\draw (31) to (21.center);
		\draw (26) to (30);
		\draw (30) to (31);
		\draw (30) to (32.center);
		\draw (38) to (33.center);
		\draw (38) to (36.center);
		\draw [style=diredge] (43.center) to (44.center);
		\draw (45.center) to (46.center);
		\draw (61.center) to (57);
		\draw (57) to (58.center);
		\draw (59.center) to (52);
		\draw (52) to (51);
		\draw (51) to (60.center);
		\draw (47) to (48);
		\draw (48) to (49);
		\draw (49) to (50);
		\draw (38) to (48);
		\draw (49) to (51);
		\draw (52) to (54);
		\draw (55) to (57);
		\draw (54) to (55);
		\draw (53) to (54);
		\draw (55) to (56);
	\end{pgfonlayer}
\end{tikzpicture}$}
\hspace{2cm}
\scalebox{0.8}{$\begin{tikzpicture}
	\begin{pgfonlayer}{nodelayer}
		\node [style=white dot] (0) at (-5, 0) {};
		\node [style=none] (1) at (-6, 1) {};
		\node [style=none] (2) at (-5, 1) {};
		\node [style=none] (3) at (-4, 1) {};
		\node [style=none] (4) at (-4.5, -1) {};
		\node [style=none] (5) at (-5.5, -1) {};
		\node [style=none] (33) at (-0.5, 1.5) {};
		\node [style=none] (36) at (-0.5, -3.5) {};
		\node [style=white dot] (38) at (-0.5, 0.25) {};
		\node [style=none] (43) at (-3, 0) {};
		\node [style=none] (44) at (-2.25, 0) {};
		\node [style=none] (45) at (-3, 0.25) {};
		\node [style=none] (46) at (-3, -0.25) {};
		\node [style=gray dot] (47) at (0.5, 1) {};
		\node [style=gray dot] (48) at (0.5, 0.25) {};
		\node [style=gray dot] (49) at (0.5, -0.5) {};
		\node [style=gray dot] (50) at (0.5, -1.25) {};
		\node [style=white dot] (51) at (1.5, -0.5) {};
		\node [style=white dot] (52) at (1.5, -1.25) {};
		\node [style=gray dot] (53) at (2.5, -0.5) {};
		\node [style=gray dot] (54) at (2.5, -1.25) {};
		\node [style=gray dot] (55) at (2.5, -2) {};
		\node [style=gray dot] (56) at (2.5, -2.75) {};
		\node [style=white dot] (57) at (3.5, -2) {};
		\node [style=none] (59) at (1.5, -3.5) {};
		\node [style=none] (60) at (1.5, 1.5) {};
		\node [style=none] (61) at (3.5, 1.5) {};
		\node [style=white dot] (62) at (3.5, -3.5) {};
	\end{pgfonlayer}
	\begin{pgfonlayer}{edgelayer}
		\draw [bend right=15] (1.center) to (0);
		\draw (0) to (2.center);
		\draw [bend right=15] (0) to (3.center);
		\draw [bend right=15] (0) to (5.center);
		\draw [bend left=15] (0) to (4.center);
		\draw (38) to (33.center);
		\draw (38) to (36.center);
		\draw [style=diredge] (43.center) to (44.center);
		\draw (45.center) to (46.center);
		\draw (61.center) to (57);
		\draw (59.center) to (52);
		\draw (52) to (51);
		\draw (51) to (60.center);
		\draw (47) to (48);
		\draw (48) to (49);
		\draw (49) to (50);
		\draw (38) to (48);
		\draw (49) to (51);
		\draw (52) to (54);
		\draw (55) to (57);
		\draw (54) to (55);
		\draw (53) to (54);
		\draw (55) to (56);
		\draw (62) to (57);
	\end{pgfonlayer}
\end{tikzpicture}$}
\end{equation}

\noindent
Spiders with a different number of input and output legs are then obtained by application of input legs to $\ket{+}$ states or post-selection of output legs onto the Pauli X measurement outcome corresponding to the $\bra{+}$ effect.

\subsection{IQP ans\"atze}

This family of ans\"atze consists of instantaneous quantum polynomial (IQP) circuits.
IQP circuits constitute of one or more layers, each layer consisting of a row of Hadamard gates, followed by a ladder of parametrised controlled-Z rotations---the rotations commute, hence the name ``instantaneous''.
At the end, a final row of Hadamards is applied.
Here follows a schematic representation of one such circuit:

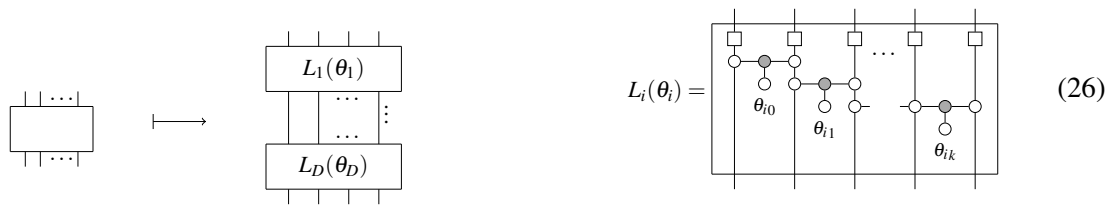
\begin{equation}
\scalebox{0.8}{$\begin{tikzpicture}
	\begin{pgfonlayer}{nodelayer}
		\node [style=none] (35) at (-3.75, 0) {};
		\node [style=none] (36) at (-8.25, 0) {};
		\node [style=none] (37) at (-8.25, 1.5) {};
		\node [style=none] (38) at (-3.75, 1.5) {};
		\node [style=none] (40) at (-7.5, 2) {};
		\node [style=none] (41) at (-6.5, 2) {};
		\node [style=none] (43) at (-5.5, 2) {};
		\node [style=none] (44) at (-4.5, 2) {};
		\node [style=none] (45) at (-7.5, 1.5) {};
		\node [style=none] (46) at (-6.5, 1.5) {};
		\node [style=none] (47) at (-5.5, 1.5) {};
		\node [style=none] (48) at (-4.5, 1.5) {};
		\node [style=none] (49) at (-7.5, 0) {};
		\node [style=none] (50) at (-6.5, 0) {};
		\node [style=none] (52) at (-4.5, 0) {};
		\node [style=none] (61) at (-7.5, 0) {};
		\node [style=none] (62) at (-6.5, 0) {};
		\node [style=none] (64) at (-4.5, 0) {};
		\node [style=none] (94) at (-5.5, -3.75) {};
		\node [style=none] (95) at (-6.5, -3.75) {};
		\node [style=none] (96) at (-7.5, -3.75) {};
		\node [style=none] (97) at (-4.5, -3.75) {};
		\node [style=none] (98) at (-3.75, -3.25) {};
		\node [style=none] (99) at (-8.25, -3.25) {};
		\node [style=none] (100) at (-8.25, -1.75) {};
		\node [style=none] (101) at (-3.75, -1.75) {};
		\node [style=none] (106) at (-7.5, -1.75) {};
		\node [style=none] (107) at (-6.5, -1.75) {};
		\node [style=none] (109) at (-4.5, -1.75) {};
		\node [style=none] (110) at (-7.5, -3.25) {};
		\node [style=none] (111) at (-6.5, -3.25) {};
		\node [style=none] (112) at (-5.5, -3.25) {};
		\node [style=none] (113) at (-4.5, -3.25) {};
		\node [style=none] (114) at (-5.5, -1.5) {$\dots$};
		\node [style=none] (115) at (-5.5, -0.25) {$\dots$};
		\node [style=none] (123) at (-4.25, -0.75) {.};
		\node [style=none] (124) at (-4.25, -1) {.};
		\node [style=none] (125) at (-4.25, -0.5) {.};
		\node [style=none] (126) at (-6, -2.5) {$L_D(\theta_D)$};
		\node [style=none] (129) at (-6, 0.75) {$L_1(\theta_1)$};
		\node [style=none] (130) at (-16.75, -0.5) {};
		\node [style=none] (131) at (-14, -0.5) {};
		\node [style=none] (132) at (-16.75, -2) {};
		\node [style=none] (133) at (-14, -2) {};
		\node [style=none] (134) at (-16.25, -0.5) {};
		\node [style=none] (135) at (-16.25, 0) {};
		\node [style=none] (136) at (-15.75, -0.5) {};
		\node [style=none] (137) at (-15.75, 0) {};
		\node [style=none] (138) at (-14.5, -0.5) {};
		\node [style=none] (139) at (-14.5, 0) {};
		\node [style=none] (140) at (-15, -0.25) {};
		\node [style=none] (141) at (-15, -0.25) {$\dots$};
		\node [style=none] (142) at (-16.25, -2.5) {};
		\node [style=none] (143) at (-16.25, -2) {};
		\node [style=none] (144) at (-15.75, -2.5) {};
		\node [style=none] (145) at (-15.75, -2) {};
		\node [style=none] (146) at (-14.5, -2.5) {};
		\node [style=none] (147) at (-14.5, -2) {};
		\node [style=none] (148) at (-15, -2.25) {};
		\node [style=none] (149) at (-15, -2.25) {$\dots$};
		\node [style=none] (150) at (-12, -0.75) {};
		\node [style=none] (151) at (-12, -1.25) {};
		\node [style=none] (152) at (-12, -1) {};
		\node [style=none] (153) at (-10.25, -1) {};
	\end{pgfonlayer}
	\begin{pgfonlayer}{edgelayer}
		\draw [style=none] (37.center) to (38.center);
		\draw [style=none] (38.center) to (35.center);
		\draw [style=none] (35.center) to (36.center);
		\draw [style=none] (36.center) to (37.center);
		\draw (40.center) to (45.center);
		\draw (41.center) to (46.center);
		\draw (43.center) to (47.center);
		\draw (44.center) to (48.center);
		\draw [style=none] (100.center) to (101.center);
		\draw [style=none] (101.center) to (98.center);
		\draw [style=none] (98.center) to (99.center);
		\draw [style=none] (99.center) to (100.center);
		\draw (113.center) to (97.center);
		\draw (94.center) to (112.center);
		\draw (111.center) to (95.center);
		\draw (96.center) to (110.center);
		\draw (130.center) to (131.center);
		\draw (133.center) to (132.center);
		\draw (130.center) to (132.center);
		\draw (133.center) to (131.center);
		\draw (135.center) to (134.center);
		\draw (136.center) to (137.center);
		\draw (139.center) to (138.center);
		\draw (143.center) to (142.center);
		\draw (144.center) to (145.center);
		\draw (147.center) to (146.center);
		\draw (150.center) to (151.center);
		\draw [style=diredge] (152.center) to (153.center);
		\draw (61.center) to (106.center);
		\draw (62.center) to (107.center);
		\draw (109.center) to (64.center);
	\end{pgfonlayer}
\end{tikzpicture}$}
\hspace{3cm}
\scalebox{0.8}{$L_i(\theta_i) = \begin{tikzpicture}
	\begin{pgfonlayer}{nodelayer}
		\node [style=none] (3) at (-2.5, 1.25) {$\dots$};
		\node [style=H box] (4) at (-7.5, 1.75) {};
		\node [style=H box] (5) at (-5.5, 1.75) {};
		\node [style=H box] (6) at (-1.5, 1.75) {};
		\node [style=white dot] (7) at (-7.5, 1) {};
		\node [style=white dot] (8) at (-5.5, 1) {};
		\node [style=white dot] (9) at (-6.5, 0.25) {};
		\node [style=gray dot] (10) at (-6.5, 1) {};
		\node [style=white dot] (11) at (-5.5, 0.25) {};
		\node [style=white dot] (12) at (-3.5, 0.25) {};
		\node [style=white dot] (13) at (-4.5, -0.5) {};
		\node [style=gray dot] (14) at (-4.5, 0.25) {};
		\node [style=white dot] (15) at (-1.5, -0.5) {};
		\node [style=none] (16) at (-2, -0.5) {};
		\node [style=none] (17) at (-3, -0.5) {};
		\node [style=white dot] (18) at (-3.5, -0.5) {};
		\node [style=none] (19) at (-1.5, -3.25) {};
		\node [style=none] (20) at (-3.5, -3.25) {};
		\node [style=none] (21) at (-5.5, -3.25) {};
		\node [style=none] (22) at (-7.5, -3.25) {};
		\node [style=none] (24) at (-6.5, -0.5) {\footnotesize ${\theta_i}_0$};
		\node [style=none] (25) at (-4.5, -1.25) {\footnotesize ${\theta_i}_1$};
		\node [style=white dot] (26) at (0.5, -0.5) {};
		\node [style=white dot] (27) at (-0.5, -1.25) {};
		\node [style=gray dot] (28) at (-0.5, -0.5) {};
		\node [style=none] (30) at (0.5, -3.25) {};
		\node [style=none] (32) at (-0.5, -2) {\footnotesize ${\theta_i}_k$};
		\node [style=H box] (33) at (0.5, 1.75) {};
		\node [style=H box] (34) at (-3.5, 1.75) {};
		\node [style=none] (35) at (1.25, -2.75) {};
		\node [style=none] (36) at (-8.25, -2.75) {};
		\node [style=none] (37) at (-8.25, 2.25) {};
		\node [style=none] (38) at (1.25, 2.25) {};
		\node [style=none] (40) at (-7.5, 2.75) {};
		\node [style=none] (41) at (-5.5, 2.75) {};
		\node [style=none] (42) at (-3.5, 2.75) {};
		\node [style=none] (43) at (-1.5, 2.75) {};
		\node [style=none] (44) at (0.5, 2.75) {};
	\end{pgfonlayer}
	\begin{pgfonlayer}{edgelayer}
		\draw (7) to (4);
		\draw (7) to (22.center);
		\draw (7) to (10);
		\draw (10) to (8);
		\draw (10) to (9);
		\draw (5) to (8);
		\draw (8) to (11);
		\draw (11) to (21.center);
		\draw (11) to (14);
		\draw (14) to (12);
		\draw (14) to (13);
		\draw (20.center) to (18);
		\draw (18) to (12);
		\draw (18) to (17.center);
		\draw (6) to (15);
		\draw (15) to (19.center);
		\draw (16.center) to (15);
		\draw (28) to (26);
		\draw (28) to (27);
		\draw (26) to (30.center);
		\draw (15) to (28);
		\draw (33) to (26);
		\draw (34) to (12);
		\draw [style=none] (37.center) to (38.center);
		\draw [style=none] (38.center) to (35.center);
		\draw [style=none] (35.center) to (36.center);
		\draw [style=none] (36.center) to (37.center);
		\draw (44.center) to (33);
		\draw (43.center) to (6);
		\draw (42.center) to (34);
		\draw (41.center) to (5);
		\draw (40.center) to (4);
	\end{pgfonlayer}
\end{tikzpicture}$}
\end{equation}

\noindent
As with the previous family, state ans\"atze are obtained by application to $\ket{0}$ states and effect ans\"atze are obtained by post-selection against $\bra{0}$ effects.
Transposition of an IQP ansatz results in another IQP ansatz with layers and rotations in reverse order.
Reversal of all inputs and outputs of an IQP ansatz results in another IQP ansatz with layers in the same order but rotations in reverse order.

Finally, as very simple example, let us see what the variational circuit for the grammatical sentence "Basil paints Dorian",
where we reshape the diagram with the {\tt bigraph} method
as is done in \cite{Zeng2016}, and we choose a $4$-dimensional noun-space and $2$-dimensional sentence-space, and use a depth-one IQP ansatz for the words:
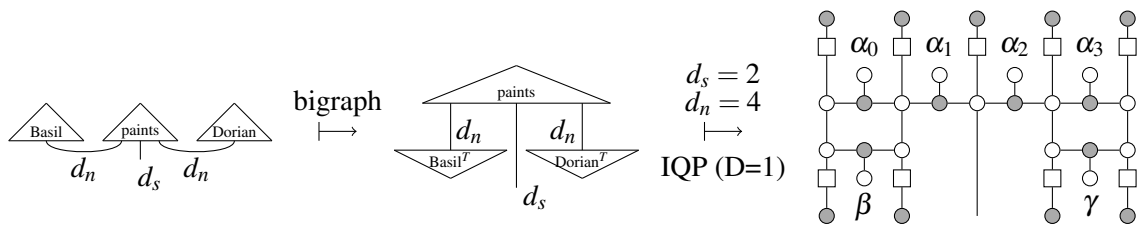
\[    \begin{tikzpicture}
	\begin{pgfonlayer}{nodelayer}
		\node [style=none] (0) at (-14.75, 1) {};
		\node [style=none] (1) at (-12.75, 1) {};
		\node [style=none] (2) at (-13.75, 2) {};
		\node [style=none] (3) at (-12.25, 1) {};
		\node [style=none] (4) at (-10.25, 1) {};
		\node [style=none] (5) at (-11.25, 2) {};
		\node [style=none] (6) at (-9.75, 1) {};
		\node [style=none] (7) at (-7.75, 1) {};
		\node [style=none] (8) at (-8.75, 2) {};
		\node [style=none] (9) at (-11.75, 1) {};
		\node [style=none] (10) at (-10.75, 1) {};
		\node [style=none] (11) at (-13.75, 1) {};
		\node [style=none] (12) at (-8.75, 1) {};
		\node [style=none] (13) at (-11.25, 1) {};
		\node [style=none] (14) at (-11.25, 1.25) {\tiny paints};
		\node [style=none] (15) at (-8.75, 1.25) {\tiny Dorian};
		\node [style=none] (16) at (-13.75, 1.25) {\tiny Basil};
		\node [style=none] (17) at (-11.25, 0.5) {};
		\node [style=none] (18) at (-12.75, 0.25) {$d_n$};
		\node [style=none] (19) at (-11, 0) {$d_s$};
		\node [style=none] (20) at (-9.75, 0.25) {$d_n$};
		\node [style=none] (21) at (-6.5, 1) {};
		\node [style=none] (22) at (-6.5, 1.25) {};
		\node [style=none] (23) at (-6.5, 1.5) {};
		\node [style=none] (24) at (-5.5, 1.25) {};
		\node [style=none] (25) at (-6, 2) {};
		\node [style=none] (26) at (-6, 2) {bigraph};
		\node [style=none] (27) at (-4.5, 0.75) {};
		\node [style=none] (28) at (-1.5, 0.75) {};
		\node [style=none] (29) at (-3, 0) {};
		\node [style=none] (30) at (-3.75, 2) {};
		\node [style=none] (31) at (1.25, 2) {};
		\node [style=none] (32) at (-1.25, 3) {};
		\node [style=none] (33) at (-1, 0.75) {};
		\node [style=none] (34) at (2, 0.75) {};
		\node [style=none] (35) at (0.5, 0) {};
		\node [style=none] (36) at (-3, 2) {};
		\node [style=none] (37) at (0.5, 2) {};
		\node [style=none] (38) at (-3, 0.75) {};
		\node [style=none] (39) at (0.5, 0.75) {};
		\node [style=none] (40) at (-1.25, 2) {};
		\node [style=none] (41) at (-1.25, 2.25) {\tiny paints};
		\node [style=none] (42) at (0.5, 0.5) {\tiny Dorian$^T$};
		\node [style=none] (43) at (-3, 0.5) {\tiny Basil$^T$};
		\node [style=none] (44) at (-1.25, -0.25) {};
		\node [style=none] (45) at (-2.5, 1.25) {$d_n$};
		\node [style=none] (46) at (-0.75, -0.5) {$d_s$};
		\node [style=none] (47) at (0, 1.25) {$d_n$};
		\node [style=none] (48) at (3.75, 1) {};
		\node [style=none] (49) at (3.75, 1.25) {};
		\node [style=none] (50) at (3.75, 1.5) {};
		\node [style=none] (51) at (4.75, 1.25) {};
		\node [style=none] (52) at (4.25, 2) {};
		\node [style=none] (53) at (4.25, 2) {$d_n=4$};
		\node [style=none] (54) at (4.25, 2.75) {$d_s=2$};
		\node [style=none] (55) at (4.25, 0.25) {IQP (D=1)};
		\node [style=white dot] (71) at (7, 2) {};
		\node [style=white dot] (72) at (9, 2) {};
		\node [style=white dot] (73) at (11, 2) {};
		\node [style=gray dot] (74) at (8, 2) {};
		\node [style=gray dot] (75) at (10, 2) {};
		\node [style=white dot] (76) at (10, 2.75) {};
		\node [style=white dot] (77) at (8, 2.75) {};
		\node [style=H box] (78) at (7, 3.5) {};
		\node [style=H box] (79) at (9, 3.5) {};
		\node [style=H box] (80) at (11, 3.5) {};
		\node [style=gray dot] (81) at (7, 4.25) {};
		\node [style=gray dot] (82) at (9, 4.25) {};
		\node [style=gray dot] (83) at (11, 4.25) {};
		\node [style=white dot] (84) at (13, 2) {};
		\node [style=white dot] (85) at (15, 2) {};
		\node [style=gray dot] (86) at (14, 2) {};
		\node [style=white dot] (87) at (14, 2.75) {};
		\node [style=H box] (88) at (13, 3.5) {};
		\node [style=H box] (89) at (15, 3.5) {};
		\node [style=gray dot] (90) at (13, 4.25) {};
		\node [style=gray dot] (91) at (15, 4.25) {};
		\node [style=gray dot] (92) at (12, 2) {};
		\node [style=white dot] (93) at (12, 2.75) {};
		\node [style=white dot] (94) at (7, 0.75) {};
		\node [style=white dot] (95) at (9, 0.75) {};
		\node [style=gray dot] (96) at (8, 0.75) {};
		\node [style=white dot] (97) at (8, 0) {};
		\node [style=H box] (98) at (7, 0) {};
		\node [style=H box] (99) at (9, 0) {};
		\node [style=gray dot] (100) at (7, -1) {};
		\node [style=gray dot] (101) at (9, -1) {};
		\node [style=white dot] (102) at (13, 0.75) {};
		\node [style=white dot] (103) at (15, 0.75) {};
		\node [style=gray dot] (104) at (14, 0.75) {};
		\node [style=white dot] (105) at (14, 0) {};
		\node [style=H box] (106) at (13, 0) {};
		\node [style=H box] (107) at (15, 0) {};
		\node [style=gray dot] (108) at (13, -1) {};
		\node [style=gray dot] (109) at (15, -1) {};
		\node [style=none] (110) at (11, -1) {};
		\node [style=none] (111) at (8, 3.5) {$\alpha_0$};
		\node [style=none] (112) at (10, 3.5) {$\alpha_1$};
		\node [style=none] (113) at (12, 3.5) {$\alpha_2$};
		\node [style=none] (115) at (14, 3.5) {$\alpha_3$};
		\node [style=none] (116) at (8, -0.75) {$\beta$};
		\node [style=none] (117) at (14, -0.75) {$\gamma$};
	\end{pgfonlayer}
	\begin{pgfonlayer}{edgelayer}
		\draw (0.center) to (1.center);
		\draw (0.center) to (2.center);
		\draw (2.center) to (1.center);
		\draw (3.center) to (4.center);
		\draw (3.center) to (5.center);
		\draw (5.center) to (4.center);
		\draw (6.center) to (7.center);
		\draw (6.center) to (8.center);
		\draw (8.center) to (7.center);
		\draw (13.center) to (17.center);
		\draw [bend right=90, looseness=0.50] (10.center) to (12.center);
		\draw [bend right=90, looseness=0.50] (11.center) to (9.center);
		\draw (23.center) to (21.center);
		\draw [style=diredge] (22.center) to (24.center);
		\draw (27.center) to (28.center);
		\draw (27.center) to (29.center);
		\draw (29.center) to (28.center);
		\draw (30.center) to (31.center);
		\draw (30.center) to (32.center);
		\draw (32.center) to (31.center);
		\draw (33.center) to (34.center);
		\draw (33.center) to (35.center);
		\draw (35.center) to (34.center);
		\draw (40.center) to (44.center);
		\draw [in=90, out=-90, looseness=0.25] (37.center) to (39.center);
		\draw [in=-90, out=90, looseness=0.75] (38.center) to (36.center);
		\draw (50.center) to (48.center);
		\draw [style=diredge] (49.center) to (51.center);
		\draw (78) to (71);
		\draw (71) to (74);
		\draw (74) to (77);
		\draw (74) to (72);
		\draw (72) to (79);
		\draw (72) to (75);
		\draw (75) to (76);
		\draw (75) to (73);
		\draw (73) to (80);
		\draw (81) to (78);
		\draw (82) to (79);
		\draw (83) to (80);
		\draw (88) to (84);
		\draw (84) to (86);
		\draw (86) to (87);
		\draw (86) to (85);
		\draw (85) to (89);
		\draw (73) to (84);
		\draw (90) to (88);
		\draw (91) to (89);
		\draw (92) to (93);
		\draw (98) to (94);
		\draw (94) to (96);
		\draw (96) to (97);
		\draw (96) to (95);
		\draw (95) to (99);
		\draw (100) to (98);
		\draw (101) to (99);
		\draw (106) to (102);
		\draw (102) to (104);
		\draw (104) to (105);
		\draw (104) to (103);
		\draw (103) to (107);
		\draw (108) to (106);
		\draw (109) to (107);
		\draw (73) to (110.center);
		\draw (71) to (94);
		\draw (72) to (95);
		\draw (84) to (102);
		\draw (85) to (103);
	\end{pgfonlayer}
\end{tikzpicture} \]
The $\alpha$s parameterise the verb "paints" and
$\beta$ and $\gamma$ parameterise each of the nounds "Basil" and "Dorian"
as angles of the controlled-Z rotations.
Note the post-selection needed to prepare the $1$-qubit state 
carrying the semantics of the sentence.

\section{Future Work}

In this work, we have described a pipeline for the implementation of NLP tasks on near-term quantum devices, by compositional translation of lexical structures to parametrised quantum circuits.
We have provided two methods---named \texttt{bigraph} and \texttt{snake_removal}---for optimising the resulting circuits, developed with the goal of near-term implementation on NISQ devices.

These are humble first steps in uncharted territory and much work remains to be done on the practical side of things.
Firstly, our optimisation methods are limited by the assumption that qubits be arranged linearly, and the job of making optimal use of each machine-specific arrangement,
such as for example minimizing swaps (wire crossings),
is left to the transpiler of the specific device,
such as IBM's,
or an independent compiler, such as $\mathrm{t|ket\rangle}$.
Future work will take qubit topology into consideration when minimising the number of crossings in the \texttt{bigraph} method.
Secondly, we intend to put a lot of work into benchmarking the optimisation algorithms, ansatz choices, training methods and various hyper-parameters, including an investigation of the relationship between corpus size, wire dimensionality and generalisation.
Finally, we will explore alternative quantum computing models, such as continuous-variable, adiabatic and measurement-based.

A lot also remains to be done on the theoretical side.
As an example, we note that not all linguistic phenomena are well approximated by the use of context-free grammars.
Lambek himself proposed the introduction of a ``meet'' operation, combining two or more pregroup grammars to give rise to a context-sensitive grammar~\cite{lambek2006pregroups}.
A class of grammars for natural language are called mildly context-sensitive \cite{mildlycsg}.
It will be interesting to investigate the various ways in which such a model can be mapped onto quantum circuits, to accommodate our choice of distributional semantics in $\textbf{fHilb}$.

Another interesting question concerns the incorporation of mixed behaviour in the semantics themselves, moving from $\textbf{fHilb}$ to the operator model of quantum theory.
Density matrices have already been used in the literature to model entailment and ambiguity \cite{Sadrzadeh2018,MarthaDot} and they can be practically realised with quantum circuits by incorporating measurements and controlled operations, with polynomial overhead.
Indeed, parsing has been assumed as input to our pipeline,
but the general problem of parsing is not trivial.
Efficiently dealing with ambiguity is crucial,
if one wishes to gain a benefit from the competition between
potential quantum advantage of any kind
versus classical overheads arising from for example
derivational or parsing ambiguity
or postprocessing of measurement outcomes from a quantum computer.

\section*{Acknowledgements}

KM and SG would like to acknowledge useful and interesting discussions with Vojt\v{e}ch Havl\'{i}\v{c}ek and Antonin Delpeuch.
KM is supported by a ih857Research Fellowship by the Royal Commission for the Exhibition of 1851 ({\tt royalcommission1851.org}).
All diagrams were drawn with TikZiT ({\tt tikzit.github.io}).
KM, GDF, AT and BC would like to acknowledge financial support from Cambridge Quantum Computing Ltd.
SG and NC would like to acknowledge financial support from Hashberg Ltd.

\bibliographystyle{eptcs}
\bibliography{refs}

\end{document}